\pdfoutput=1
\documentclass{article}

\usepackage{graphicx}
\usepackage{caption}
\usepackage{subcaption}
\usepackage{float}

\usepackage{amsmath, amsfonts, amsthm, amssymb}
\usepackage[utf8]{inputenc}
\usepackage{geometry}[margin=1in]   
\usepackage[backend=biber, style=numeric, sorting=nty]{biblatex}
\usepackage{url}
\usepackage{hyperref}
\usepackage[bb=boondox]{mathalfa}   
\usepackage{authblk}
\usepackage[symbol]{footmisc}       

\newtheorem{theorem}{Theorem}

\newtheorem{assumption}{Assumption}

\usepackage{tabulary}
\usepackage{booktabs}
\usepackage[table]{xcolor}

\usepackage[ruled, lined, linesnumbered, commentsnumbered, longend]{algorithm2e}
\SetKwProg{Init}{Initialization}{}{}

\title{\texttt{DDE-Find}: Learning Delay Differential Equations from Noisy, Limited Data}
\author[a]{Robert Stephany\thanks{Corresponding Author. 
Email address: \href{mailto:rrs254@cornell.edu}{\texttt{rrs254@cornell.edu}}}$^,$}
\date{}

\affil[a]{\it Center for Applied Mathematics, 657 Frank H.T. Rhodes Hall, Cornell University, Ithaca, NY 14853, United States}

\begin{document}
\maketitle
\begin{abstract}
\noindent 
Delay Differential Equations (DDEs) are a class of differential equations that can model diverse scientific phenomena.
However, identifying the parameters, especially the time delay, that make a DDE's predictions match experimental results can be challenging.
We introduce \texttt{DDE-Find}, a data-driven framework for learning a DDE's parameters, time delay, and initial condition function.
\texttt{DDE-Find} uses an adjoint-based approach to efficiently compute the gradient of a loss function with respect to the model parameters.
We motivate and rigorously prove an expression for the gradients of the loss using the adjoint.
\texttt{DDE-Find} builds upon recent developments in learning DDEs from data and delivers the first complete framework for learning DDEs from data.
Through a series of numerical experiments, we demonstrate that \texttt{DDE-Find} can learn DDEs from noisy, limited data.
\end{abstract}

\section{Introduction}
\label{Sec:Introduction}

Scientific progress, especially in the physical sciences, relies on finding mathematical models --- often as differential equations --- to describe the time evolution of physical systems.
Historically, scientists discovered these models using a \emph{first-principles} approach: distilling results from controlled experiments into axiomatic first principles, from which they derive predictive mathematical models.
This approach has yielded myriad scientific models, from Newtonian mechanics to general relativity.
Despite its success, the first principles approach does have its limitations.
For one, many physical systems lack predictive models \cite{abreu2019mortality} \cite{amor2020transient}.
Further, even if scientist propose a predictive model, the model generally has a collection of unknown parameters (these could be coefficients in the differential equation, initial conditions, etc.).
In such instances, we need to be able to search for a set of parameters for which the model's predictions match experimental results.

\bigskip 

\emph{Neural Ordinary Differential Equations} (NODEs) offer one potential solution
 \cite{chen2018neural}.
NODEs begin with a collection of inputs and targets.
They assume that the map from input to target can be described using an ordinary differential equation \emph{initial value problem} (IVP) where the input, $x_0$, is the IVP's initial condition and the target value is the IVP's solution at a later time, $T$.
The IVP contains a set of unknown parameters.
The NODE algorithm trains these parameters such that the IVP approximately maps inputs to target values. 
Though NODEs were not originally proposed as a tool for scientific discovery, they hold tremendous potential for such applications. 

\bigskip 

Delay Differential Equations (DDEs) are a class of differential equations where the time derivative of the solution, $x$, depends on the system's current state, $x(t)$, and the state at a previous or \emph{delayed} time, $x(t - \tau)$.
Though they have generally received less attention than their ordinary or partial differential equation counterparts, DDEs represent an important modeling tool in scientific disciplines.
For instance, DDEs can model HIV infections, where the delay term models the time delay between when a T-cell first contacts the HIV and when it begins producing new viruses \cite{nelson2000model}.
DDEs can also model computer networks, where the delay term arises from the fact that transmitting information takes time \cite{yu2004lmi}.
Finally, in population dynamics, DDEs can model the time it takes for members of a species to mature before they start reproducing \cite{kuang1993delay}.
In each instance, however, to reconcile the DDE model with experimental data, we need tools that learn a collection of parameters in the model such that the model's predictions match experimental results.

\bigskip 

Critically, any tool designed to discover scientific models must be able to work with scientific data.
Scientific data sets can be limited (sometimes only a few hundred data points) and generally contain noise.
Therefore, any tool that aims to match scientific laws to experimental data must be able to work with limited, noisy data.

\bigskip 

{\bf Contributions:} This paper presents a novel algorithm, \texttt{DDE-Find}, that can learn a DDE model from noisy, limited data.
Specifically, \texttt{DDE-Find} uses an approach derived from the NODE framework to simultaneously learn the initial condition function, model parameters, and delay in a DDE model.
\texttt{DDE-Find} can learn both linear and non-linear models.
We demonstrate \texttt{DDE-Find}'s efficacy through a series of numerical experiments that reveal that \texttt{DDE-Find} performs well even when the data set contains considerable noise. 

\bigskip 

{\bf Outline:} The rest of this paper is organized as follows:
First, in section \ref{Sec:Problem}, we formally state the problem \texttt{DDE-Find} aims to solve, including all assumptions we make about the available data and the underlying DDE model.
Next, in section \ref{Sec:Adjoint}, we introduce adjoint approaches and state a critical theorem that underpins the entire \texttt{DDE-Find} algorithm (we prove this theorem in section \ref{Sec:Proof}).
In section \ref{Sec:Method}, we introduce the \texttt{DDE-Find} algorithm.
Then, in section \ref{Sec:Experiments}, we demonstrate that \texttt{DDE-Find} can identify several DDEs from noisy, limited measurements.
We then discuss some other aspects, quirks, and properties of the \texttt{DDE-Find} algorithm in section \ref{Sec:Discussion}.
Finally, we state some concluding remarks in section \ref{Sec:Conclusion}.
\section{Problem Statement}
\label{Sec:Problem}

Consider the following delay differential equation \emph{initial value problem} (IVP):

\begin{equation}
\begin{aligned}
    \dot{x}(t) &= F\left(x(t), x(t - \tau), \tau, t, \theta\right) && t \in (0, T) \\ 
    x(t) &= X_0(t, \phi) && t \in [-\tau, 0]
\end{aligned}
\label{eq:IVP}
\end{equation}

Here, $T > 0$ and $\tau > 0$ are the final time and delay, respectively.
Further, $F : \mathbb{R}^d \times \mathbb{R}^d \times (0, \infty) \times \mathbb{R} \times \mathbb{R}^p \to \mathbb{R}^d$ is a continuously differentiable function (it has continuous partial derivatives with respect to each component of each of its arguments).
$F$ defines the right-hand side of the dynamics governing $x$.
Finally, $X_0 : [-\tau, 0] \times \mathbb{R}^{q} \to \mathbb{R}^d$ is a function that is continuously differentiable with respect to its second argument (it has continuous partial derivatives with respect to each component of $\phi$). 
We do not assume $X_0$ is differentiable with respect to its first argument, $t$.
$X_0$ represents the initial condition function of the IVP.

\bigskip 

Notice that \eqref{eq:IVP} has a similar form to an IVP for an ordinary differential equation, with a few key differences.
First, the time derivative of $x$ depends on the current state, $x(t)$, and the state at a delayed time, $x(t - \tau)$.
Because of this, we need to know $x(t)$ for $t \in [-\tau, 0]$ to compute $\dot{x}(t)$ for $t \in [0, \tau]$. 
For this reason, the initial condition for a DDE is a function (defined on $[-\tau, 0]$) rather than a single value (as is the case in an ODE).

\bigskip 

Let $\{ \tilde{x}(t_i) \}_{i = 1}^{N_{Data}}$ be a collection of $N_{Data} \in \mathbb{N}$ noisy measurements of a function, $x : [-\tau, T] \to \mathbb{R}^d$, which solves equation \eqref{eq:IVP}.
By a solution, we mean that there are specific $\tau > 0$, $\phi \in \mathbb{R}^q$ and $\theta \in \mathbb{R}^p$ such that $x$ is the solution to equation \eqref{eq:IVP} when the IVP uses those parameters.
By noisy measurements, we mean there is a probability space $(\Theta, \Sigma, P)$ \cite{durrett2019probability}, a collection of random variables, $\varepsilon_1, \ldots, \varepsilon_{N_{Data}} : \Omega \to \mathbb{R}^d$, and some $\omega \in \Omega$ such that for each $i \in \{ 1, 2, \ldots, N_{Data} \}$,
$$\tilde{x}(t_i) - x(t_i) = \varepsilon_{i}(\omega).$$
We assume we know the times, $t_1, \ldots, t_{N_{Data}} \in [0, T]$, but not the true solution, $x$.
Further, for simplicity, we assume that $0 = t_1 < t_2 < \cdots < t_{N_{Data}} = T$.
Throughout this paper, we assume that $\varepsilon_i \sim N(0, \sigma^2 )$, for some $\sigma > 0$. 
That is, we assume the noise is Gaussian with mean zero and standard deviation $\sigma$.
Finally, we assume that the noise random variables, $\varepsilon_1, \ldots, \varepsilon_{N_{Data}}$ are independent and identically distributed.
We refer to the function, $x$, of which we have noisy measurements as the \emph{true trajectory}.
Our goal is to use the noisy measurements of the true trajectory, $x$, to recover $\tau$, $\phi$, and $\theta$ in the IVP that $x$ satisfies.

\bigskip 

Table \ref{Table:Problem:Notation} lists the notation we introduced in this section.

\begin{table}[hbt]
    \centering 
    \rowcolors{2}{white}{cyan!10}
    
    \begin{tabulary}{1.0\linewidth}{p{1.5cm}L}
        \toprule[0.3ex]
        \textbf{Notation} & \textbf{Meaning} \\
        \midrule[0.1ex]

        $C^k(A, B)$ & The set of functions $f$ from $A$ to $B$, where $A$ is an open subset of a Euclidean space and $B$ is a (possibly different) Euclidean space, such that each component function of $f$ has continuous partial derivatives of order $\leq k$ on $A$. If $f \in C^k(A, B)$, we sometimes say that $f$ is of class $C^k$ on $A$. \\ 
        \addlinespace[0.4em]
        $d$ & The dimension of the space in which the dynamics occurs. \\
        \addlinespace[0.4em]
        $T$ & The final time. See equation \eqref{eq:IVP}. \\
        \addlinespace[0.4em]
        $x$ & The solution to the IVP \eqref{eq:IVP}. \\
        \addlinespace[0.4em]
        $\tau$ & The time delay in the DDE. See equation \eqref{eq:IVP}. \\
        \addlinespace[0.4em]
        $\theta$ & An element of $\mathbb{R}^p$ representing the parameters of $F$, the right-hand side of the DDE \eqref{eq:IVP}. \\
        \addlinespace[0.4em]
        $F$ & A function on $\mathbb{R}^d \times \mathbb{R}^d \times (0, \infty) \times \mathbb{R} \times \mathbb{R}^p$ that defines the right-hand side of the DDE \eqref{eq:IVP}. We assume this function is of class $C^1$. \\
        \addlinespace[0.4em]
        $\phi$ & An element of $\mathbb{R}^q$ representing the parameters of $X_0$, the initial condition function of the DDE \eqref{eq:IVP}. \\
        \addlinespace[0.4em]
        $X_0$ & A function on $[0, \tau) \times \mathbb{R}^q$ defining the initial condition function of the DDE \eqref{eq:IVP}. We assume that this function has continuous partial derivatives with respect to each component of its second argument ($\phi$). \\
        \addlinespace[0.4em]
        $\tilde{x}(t_i)$ & A noisy measurement of the true trajectory, the solution to equation \eqref{eq:IVP} whose parameters we want to recover, at time $t_i \in [0, T]$. \\
        \addlinespace[0.4em]
        $\varepsilon_i$ & A random variable representing the noise in the measurement $\tilde{x}(t_i)$. Specifically, we assume each $\varepsilon_i$ is defined on a probability space $(\Omega, \Sigma, P)$ and there is some $\omega \in \Omega$ such that $\tilde{x}(t_i) - x(t_i) = \varepsilon_i(\omega)$. We assume that $\varepsilon_i \sim N(0, \sigma^2)$. \\
        \bottomrule[0.3ex]
    \end{tabulary}
    
    \caption{The notation and terminology of section (\ref{Sec:Problem})} 
    \label{Table:Problem:Notation}
\end{table}
\section{The Adjoint Equation}
\label{Sec:Adjoint}

To recover the parameters --- $\tau, \theta, \phi$ --- from the noisy measurements of $x$, we search for a set of parameters such that if we solve the IVP, equation \eqref{eq:IVP}, using those parameters, the corresponding solution closely matches the noisy measurements of $x$.
We begin with an initial guess for the parameters and then iteratively update our guess until we find a suitable combination.
We refer to the trajectory we get when we solve the IVP, equation \eqref{eq:IVP}, using our current guess for the parameters as the \emph{predicted trajectory}.
Thus, our approach is to find a set of parameters whose corresponding predicted trajectory closely matches the noisy measurements of $x$ (and, hopefully, matches the true trajectory).

\bigskip 

We use a specially designed loss function to quantify how \emph{closely} a predicted trajectory matches the data. 
We conjecture that a set of parameters that engenders a small loss should be a good approximation of the corresponding parameters in the IVP that generated $x$.
Thus, we aim to find a minimizer of the loss function, which we accomplish using a gradient descent approach.
However, to use gradient descent, the map from the parameters to the loss must be differentiable.
This fact, while true under suitable assumptions (see theorem \ref{Theorem:1}), is by no means obvious.
Because the parameters change how $x$ evolves, there is no way to express the map from parameters to loss as a discrete sequence of elementary operations.
In a sense, the map from parameters to loss acts like a network with infinite layers \cite{chen2018neural}.
Thus, unlike conventional neural networks, we can not justify the differentiability of the map from parameters to loss by repeatedly applying the chain rule.
The structure of our approach also means that the standard approach to back-propagation using reverse-mode automatic differentiation \cite{baydin2018automatic} does not work.
We overcome these challenges using an approach based on the adjoint state method \cite{pontryagin2018mathematical} \cite{cao2003adjoint}.

\bigskip 

In this section, we define a loss function that compares the predicted trajectory with an interpolation of the noisy measurements.
We then use an adjoint approach to justify that the map from parameters to loss is differentiable.
Finally, we express the gradient of the loss with respect to the parameters using an \emph{adjoint equation}.

\subsection{Loss}
\label{Sec:Adjoint:Losss}

In this paper, we consider the following general class of loss functions:
\begin{equation}
    \mathcal{L}(x) = \int_{0}^{T} \ell(x(t))\ dt + G(x(T)) \label{eq:loss}
\end{equation}
Thus, our loss function consists of two pieces: a running loss, $\int_{0}^{T} \ell(x(t))\ dt$, which accumulates loss over the entire problem domain $[0, T]$, and a terminal loss, $G(x(T))$, which accumulates loss at the end.
Here, $\ell : \mathbb{R}^d \to \mathbb{R}$ and $G : \mathbb{R}^d \to \mathbb{R}$ are user-selected, continuously differentiable functions.

\subsection{The Adjoint Equation}
\label{Sec:Adjoint:Adjoint}

The loss function, equation \eqref{eq:loss}, implicitly depends on the parameters.
Changing $\theta, \tau, \phi$ changes the solution, $x$.
Specifically, for each $t \in [0, T]$, there is a map that sends the parameters to $x(t)$, the solution of the IVP at time $t$. 
However, there's no easy way to express that map as a composition of a sequence of elementary operations.
Specifically, to get $x(t)$, we have to solve the IVP up to time $t$.
To highlight why this creates some issues, consider naively differentiating (assuming everything is differentiable and smooth enough to interchange integration and differentiation) $\mathcal{L}$ with respect to $\tau$:
\begin{equation}
\label{eq:Loss:Grad:Tau:Naive}
    \frac{\partial \mathcal{L}}{\partial \tau} = \int_{0}^{T} \Big( \nabla_{x} \ell\left(x(t)\right) \Big) \partial_{\tau} x(t)\ dt + \Big( \nabla_{x} G\left(x(T) \right) \Big) \partial_{\tau} x(T).
\end{equation}
To evaluate this expression, we need to know $\partial_{\tau} x(t)$ for each $t \in [0, T]$.
Since $x$ is the result of a DDE solve, and because the DDE depends on the parameters, it is reasonable to consider the differentiability of the map that sends the parameters to $x(t)$.
However, it's not entirely clear how to \emph{differentiate through the IVP solve}.
Remarkably, however, we can resolve this issue by introducing an \emph{adjoint}.

\bigskip

Before delving into our adjoint approach, we need to establish a few assumptions.
Throughout this paper, we make several assumptions about the smoothness of the functions in equation \eqref{eq:IVP} and \eqref{eq:loss}.
Specifically, we assume the following:
\begin{assumption}
\label{assumption:FGell}
    The function $F : \mathbb{R}^d \times \mathbb{R}^d \times (0, \infty) \times \mathbb{R} \times \mathbb{R}^p \to \mathbb{R}^d$ in equation \eqref{eq:IVP} is of class $C^1$. 
    Further, the functions $\ell : \mathbb{R}^d \to \mathbb{R}$, and $G : \mathbb{R}^d \to \mathbb{R}$ in equation \eqref{eq:loss} are of class $C^1$.
\end{assumption}

\begin{assumption}
\label{assumption:X0}
    Each component function of $X_0 : [0, \tau] \times \mathbb{R}^q \to \mathbb{R}^d$ (defined in \eqref{eq:IVP}) has continuous partial derivatives with respect to each component of $\phi$.
    Thus, for each $t \in [-\tau, 0]$, the restriction $X_0(t, \bullet ) : \mathbb{R}^q \to \mathbb{R}^d$ is of class $C^1$.
\end{assumption}

\begin{assumption}
\label{assumption:x}
    For $t \in [0, T]$, $\theta \in \mathbb{R}^p$, $\tau > 0$, and $\phi \in \mathbb{R}^q$, the map $(t, \theta, \tau, \phi) \to x(t)$ (where $x$ is the solution to equation \eqref{eq:IVP} when $F$'s parameters are $\theta$, the delay is $\tau$, and $X_0$'s parameters are $\phi$) is of class $C^2$.
\end{assumption}

\bigskip 

Let $\partial_x F$, $\partial_y F$, $\partial_\tau F$, and $\partial_\theta F$ denote the (Frechet) derivative \cite{munkres2018analysis} of $F$ with respect to its first, second, third, and fifth arguments, respectively .
Similarly, let $\partial_\phi X_0$ denote the Frechet derivative of $X_0$ with respect to its second argument.
Our main result is the following theorem, which gives us a convenient expression for the gradients of $\mathcal{L}$ in terms of the \emph{adjoint}, $\lambda : [0, T] \to \mathbb{R}^d$.

\bigskip

\begin{theorem}
\label{Theorem:1}
    Let $T, \tau > 0$, $\theta \in \mathbb{R}^{p}$, and $\phi \in \mathbb{R}^{q}$.
    Let $x : [-\tau, T] \to \mathbb{R}^d$ solve the initial value problem in equation \eqref{eq:IVP}.
    Further, suppose that assumptions \ref{assumption:FGell}, \ref{assumption:X0}, and \ref{assumption:x} hold. 
    Let $\lambda: [0, T] \rightarrow \mathbb{R}^n$ satisfy the following:
    \begin{equation}
    \label{eq:adjoint}
    \begin{aligned}
        \dot{\lambda}(t) & = \nabla_x \ell \left( x(t) \right) - \left[ \partial_x F\left(x(t), x(t - \tau), \tau, t, \theta\right) \right]^T \lambda(t) \\
        &\quad - \mathbb{1}_{t < T - \tau}(t) \left[ \partial_y F \left( x(t + \tau), x(t), \tau, t + \tau, \theta\right) \right]^T \lambda\left(t + \tau \right) \qquad \ t \in [0, T]  \\
        \lambda(T) &= -\nabla_x G \left(x(T)\right)
    \end{aligned}
    \end{equation}
    Then, the derivatives of the loss function, $\mathcal{L}$, with respect to $\theta, \tau$ and $\phi$ are:
    \begin{equation}
    \begin{aligned}
    \label{eq:gradients}
        \nabla_{\theta} \mathcal{L}\left( x \right) =& - \int_0^T  \left[ \partial_{\theta} F (x(t), x(t - \tau), \tau, t, \theta) \right]^T \lambda(t) \ dt \\
        \frac{\partial}{\partial\tau} \mathcal{L}\left( x \right) =& \int_0^{T - \tau} \left\langle \left[ \partial_y F (x(t + \tau), x(t), \tau, t + \tau, \theta) \right]^T \lambda(t + \tau),\ \dot{x}(t) \right\rangle\ dt \\
        -& \int_{0}^{T} \left\langle \partial_{\tau} F(x(t), x(t - \tau), \tau, t, \theta),\ \lambda(t) \right\rangle \ dt \\
        \nabla_{\phi} \mathcal{L}\left( x \right) =& - \left[ \partial_{\phi} X_0(0, \phi) \right]^T \lambda(0) \\
        &- \int_{0}^{\tau} \left[ \partial_\phi X_0(t - \tau, \phi) \right]^T \left[\partial_{y} F\left( x(t), X_0(t - \tau, \phi), \tau, t, \theta \right)\right]^T \lambda(t) \ dt
    \end{aligned}
    \end{equation}
\end{theorem}
\emph{Proof:} See Appendix \ref{Sec:Proof}.

\bigskip 

In equation \eqref{eq:adjoint}, $\mathbb{1}_{t < T - \tau}$ is the indicator function of the set $(-\infty, T - \tau)$.
Throughout this paper, we refer to equation \eqref{eq:adjoint} as the \emph{adjoint equation}.
Likewise, we refer to the function, $\lambda$, which solves the adjoint equation as the \emph{adjoint}.

\bigskip 

The adjoint equation is remarkable for several reasons.
For one, consider the gradient of $\mathcal{L}$ with respect to $\tau$.
Earlier in this section, we naively attempted to compute this gradient (see equation \eqref{eq:Loss:Grad:Tau:Naive}).
To evaluate that expression, we needed to compute $\partial_{\tau} x(t)$ for each $t \in [0, T]$; it's not entirely clear how to do this.
Finding $\partial_{\tau} x(t)$ is possible, though somewhat impractical.
We briefly address this in section \ref{Sec:Discussion:Adjoint}.
By contrast, the expression for $\partial \mathcal{L} / \partial \tau$ in equation \eqref{eq:gradients} depends solely on quantities we can compute so long as we know $x$ and $\lambda$.
By inspection, if we know $x$ on $[-\tau, T]$, we can solve the adjoint equation on $[0, T]$ backward in time.
This observation suggests the following approach: First, solve the IVP, equation \eqref{eq:IVP}, to get $x$ on $[0, T]$.
Second, solve the adjoint equation, equation \eqref{eq:adjoint}, to get $\lambda$ on $[0, T]$.
We can then directly compute $\partial \mathcal{L} / \partial \tau$ using $x$ and $\lambda$.
This three-step approach forms the basis of how we learn the parameters from the data, which we describe in section \ref{Sec:Method}.
Presently, however, the important takeaway is that the adjoint gives us a viable path to computing the gradients.

\bigskip

The adjoint approach is also computationally efficient.
Since the adjoint equation takes place in the same space as the IVP for $x$, equation \eqref{eq:IVP}, solving the adjoint equation is no more difficult than solving the IVP for $x$.
Thus, the computational complexity of solving the adjoint equation and computing the gradients of the loss is the same as solving the IVP for $x$ and evaluating the loss.

\bigskip 

Table \ref{Table:Adjoint:Notation} lists the notation we introduced in this section.

\begin{table}[hbt]
    \centering 
    \rowcolors{2}{cyan!10}{white}
    
    \begin{tabulary}{1.0\linewidth}{p{1.5cm}L}
        \toprule[0.3ex]
        \textbf{Notation} & \textbf{Meaning} \\
        \midrule[0.1ex]
        
        $\mathcal{L}$ & The loss function. See equation \eqref{eq:loss}. \\
        \addlinespace[0.4em]
        $\ell$ & A continuously differentiable function from $\mathbb{R}^d$ to $\mathbb{R}$ representing the running or integral loss. See equation \eqref{eq:loss}. \\
        \addlinespace[0.4em]
        $G$ & A continuously differentiable function from $\mathbb{R}^d$ to $\mathbb{R}$. $G$ represents the terminal loss. See equation \eqref{eq:loss}. \\
        \addlinespace[0.4em]
        $\lambda$ & The adjoint; a function from $\mathbb{R} \to \mathbb{R}^d$ which solves equation \eqref{eq:adjoint}. \\
        \bottomrule[0.3ex]
    \end{tabulary}
    
    \caption{The notation and terminology of section (\ref{Sec:Adjoint})} 
    \label{Table:Adjoint:Notation}
\end{table}
\section{Methodology}
\label{Sec:Method}

In this section, we describe the \texttt{DDE-Find} algorithm, which uses the noisy measurements, $\{ \tilde{x}(t_i) \}_{i = 1}^{N_{Data}}$, of the true solution to learn the parameters, $\theta, \tau, \phi$ in the IVP that the true solution satisfies.
\texttt{DDE-Find} uses trainable models for $F$ and $X_0$ and updates their parameters using an iterative process with three steps per iteration: First, it numerically solves the IVP, equation \eqref{eq:IVP}, using our models' current parameter values.
Next, \texttt{DDE-Find} solves the adjoint equation, equation \eqref{eq:adjoint}, backward in time.
Finally, \texttt{DDE-Find} computes the gradients of $\mathcal{L}$ using equation \eqref{eq:gradients} and then uses these gradients to update the model's parameters.
In this section, we describe these steps and the \texttt{DDE-Find} algorithm in greater depth.
We implemented \texttt{DDE-Find} in \texttt{pytorch} \cite{paszke2019pytorch}.
Our repository is open-source and available at \url{https://github.com/punkduckable/DDE-Find}.

\bigskip 

The models for $F$ and $X_0$ can be any parameterized models that satisfy assumptions \ref{assumption:FGell} and \ref{assumption:X0}.
Thus, most \texttt{torch.nn.Module} objects can serve as models for $F$ and $X_0$. 
These models can be neural networks, structured models with unknown parameters, or some combination of the two.

\subsection{\texorpdfstring{Initializing $F$ and $X_0$}{}}
\label{Sec:Method:Initialization}

Before training, \texttt{DDE-Find} initializes itself.
Initialization consists of two steps.
First, \texttt{DDE-Find} selects an initial set of values for $\theta$, $\tau$, $\phi$. 
If the practitioner has reasonable guesses for the value of any of these parameters, they can initialize \texttt{DDE-Find} using those guesses.
Otherwise, \texttt{DDE-Find} begins with random guesses for each parameter.
Second, \texttt{DDE-Find} interpolates the noisy measurements of the true trajectory, $\{ \tilde{x}(t_i) \}_{i = 1}^{N_{Data}}$, using cubic-splines.
Let $\tilde{x} : [0, T] \to \mathbb{R}^d$ denote that interpolation.
We refer to $\tilde{x}$ as the \emph{target trajectory}.
Since the data is noisy, $\tilde{x}$ is a noisy approximation of $x$. 
Crucially, however, the interpolation allows us to approximate the true trajectory at arbitrary points in $[0, T]$, which enables us to evaluate the running loss, $\int_{0}^{T} \ell(x(t))\ dt$.

\subsection{\texorpdfstring{The Forward Pass to Compute $\mathcal{L}$}{}}
\label{Sec:Method:Forward_Pass}

After initialization, \texttt{DDE-Find} uses an iterative process to learn the $\theta$, $\tau$, and $\phi$ in the DDE that the true solution, $x$, satisfies.
Specifically, it finds a $(\theta, \tau, \phi) \in \mathbb{R}^p \times (0, \infty) \times \mathbb{R}^q$ such that if we solve the initial value problem in equation \eqref{eq:IVP} using these parameters, the resulting trajectory closely matches the true trajectory.
To do this, \texttt{DDE-Find} trains the models for $F$ and $X_0$ for $N_{Epochs} \in \mathbb{N}$ epochs.
Each epoch, we solve the IVP, equation \eqref{eq:IVP}, to obtain the \emph{predicted solution} using the model's current guesses for $\theta$, $\tau$, and $\phi$.
We refer to this as the \emph{forward pass}.
\texttt{DDE-Find} uses a second-order Runge-Kutta method to solve the IVP forward in time and obtain a discrete numerical approximation of the predicted trajectory.
For the IVP solve, we set the time step, $\Delta t$, such that there is some $N_{\tau} \in \mathbb{N}$ for which $\tau = N_{\tau} \Delta t$.
Here, $N_{\tau}$ is a user-defined hyperparameter.
Notice that this approach does not place any restrictions on $\tau$; we compute $\Delta t$ from $\tau$, not the other way around.
We chose this approach because it simplifies evaluating $x(t - \tau)$; depending on the value of $t$, we either use the initial condition or the numerical solution from $N_{\tau}$ time steps ago.
We discuss this choice further in section \ref{Sec:Discussion:Ntau}.
One downside of this approach is that $T$ is not an integer multiple of $\Delta t$.
This property complicates solving the DDE on $[0, T]$ using a fixed step size.
To remedy this mismatch, we instead solve the DDE up to $t = N_{step} \Delta t$, where $N_{step} = \min\{ n \in \mathbb{N} : n \Delta t >= T \}$. 

\bigskip

After finding a discrete approximation of the predicted trajectory, \texttt{DDE-Find} evaluates the loss.
Unless otherwise stated, we use the $L^2$ norm of the difference between the predicted and target trajectories for $\ell$ and $G$. 
That is,
\begin{equation*}
\begin{aligned}
    \ell\left(x(t)\right) &= \left\| x(t) - \tilde{x}(t) \right\|_2^2 \\
    G\left( x(T) \right) &= \left\| x(T) - \tilde{x}(T) \right\|_2^2.
\end{aligned}
\end{equation*}
Where $\tilde{x}$ is the target trajectory, the interpolation of the noisy measurements, $\{ \tilde{x}(t_i) \}_{i = 1}^{N_{Data}}$.
\texttt{DDE-Find} uses the trapezoidal rule to compute the a discrete approximation of the running loss.
The fact that our DDE solve only finds the predicted solution at discrete steps (and that, in general, $T$ is not one of those steps) somewhat complicates this process.
However, this complication does not cause any real trouble when optimizing the parameters.
We discuss these points in greater detail in section \ref{Sec:Discussion:Loss}.

\subsection{\texorpdfstring{The Backward Pass to Compute $\nabla\mathcal{L}$}{}}
\label{Sec:Method:Backward_Pass}

After completing the forward pass and evaluating the loss, \texttt{DDE-Find} begins the backward pass to compute the gradients of $\mathcal{L}$.
\texttt{DDE-Find} begins by computing the adjoint, $\lambda$.
To do this, \texttt{DDE-Find} solves the adjoint equation, equation \eqref{eq:adjoint}, backward in time. 
As in the forward, pass, we use a second-order Runge-Kutta solve to find a numerical approximation to $\lambda$. 
For this solve, we also use the time step $\Delta t = \tau/N_{\tau}$.
This choice means that we solve for $\lambda$ at times $T, T - \Delta t, T - 2 \Delta t, \ldots, T - N_{step} \Delta t$.
Since the adjoint equation involves the predicted and target trajectories, $x$ and $\tilde{x}$, solving for $\lambda$ requires knowing both trajectories at $t = T, T - \Delta t, \ldots, T - N_{step} \Delta t$. 
In general, these times do not correspond to when we solved for the predicted solution in the forward pass.
To remedy this, we use a cubic-spline interpolation of the predicted and target trajectories during the adjoint solve.
These interpolations allow us to approximately evaluate the predicted and target trajectories at arbitrary elements of $[0, T]$.
We use the interpolation of the predicted trajectory at $t = T$ to evaluate $\nabla_x G(x(T))$ and initialize the adjoint, $\lambda$.
We also use the interpolations to evaluate the forward and predicted solutions at $t = T - k \Delta t$, $k \in \{ 0, 1, \ldots, N_{step} \}$.
The adjoint equation also involves the partial derivatives of $F$, $\ell$, and $G$.
We evaluate these partial derivatives exactly using \emph{pytorch}'s \emph{automatic differentiation} capabilities \cite{baydin2018automatic} \cite{paszke2017automatic}.

\bigskip 

We run the DDE solver for $N_{step}$ steps when solving the adjoint equation.
Since $N_{step} \Delta t > T$, the final step, $T - N_{step} \Delta t$, is non-positive. 
In particular, we also have $N_{step} = \min\{ n : T - n \Delta t \leq 0 \}$. 

\bigskip 

After solving the adjoint equation numerically, we use equation \eqref{eq:gradients} to compute the gradient of the loss with respect to $\theta$, $\tau$, and $\phi$.
To evaluate these gradients, we need to evaluate integrals whose integrands depend on the adjoint, predicted trajectory, and the target trajectory (through $\ell$ and $G$).
We approximate these integrals using a modified trapezoidal rule.
First, we evaluate the integrand at $t = T, T - \Delta t, \ldots, T - N_{step} \Delta t$ using the numeric solution for $\lambda$, the interpolation of the predicted target trajectories, and automatic differentiation.
We partition $[0, T]$ using the partition $0, T - (N_{step} - 1)\Delta t, T - (N_{step} - 2) \Delta t, \ldots, T$.
Thus, the first subinterval in this partition has a width of $T - (N_{step} - 1)\Delta t$ while the others have a width of $\Delta t$.
To approximate the integrand at $t = 0$, we use a liner interpolation of the integrand evaluated at $t = T - N_{step} \Delta t$ and $t = T - (N_{step} - 1) \Delta t$, which is a convex combination of the integrand at the two time values.
For other sub-intervals, we evaluate the trapezoidal approximation to the integrals as usual using our evaluations of the integrands.
This approach gives us an approximation of the gradients of the loss with respect to $\theta$, $\tau$, and $\phi$. 
\texttt{DDE-Find} then uses the \texttt{Adam} optimizer \cite{kingma2014adam} to update the parameters.

\bigskip 

\texttt{DDE-Find} iteratively updates $\theta$, $\tau$, and $\phi$ using this process.
We continue until either $N_{epochs}$ have passed or the loss drops below a user-specified threshold, $\mathcal{L}_{min}$, whichever comes first.
After completing, \texttt{DDE-Find} reports the learned $\theta$, $\tau$, and $\phi$ values.

\bigskip

Algorithm \ref{algorithm:DDE_Find} summarizes the \texttt{DDE-Find} algorithm.
Table \ref{Table:Adjoint:Notation} lists the notation we introduced in this section.

\begin{algorithm}[hbt!]
\caption{\texttt{DDE-Find}: Learning $\theta$, $\tau$, and $\phi$ from data}
\label{algorithm:DDE_Find}

\KwIn{Noisy measurements of the true trajectory: $\{ \tilde{x}(t_i) \}_{i = 1}^{N_{Data}}$}
\KwOut{$\theta, \tau, \phi$}
\Init{Select initial values for $\theta, \tau$, and $\phi$. Generate the target trajectory by interpolating the noisy data.}{ }
\For{$i \in \{ 1, \ldots, N_{epochs} \}$}{
    Obtain the predicted solution by solving the following IVP forward in time:
    \begin{equation*}
    \begin{aligned}
        \dot{x}(t) &= F(x(t), x(t - \tau), \tau, t, \theta) & \quad t \in [0, T] \\
        x(t) &= X_0(t, \phi) & t [-\tau, 0]
    \end{aligned}
    \end{equation*} \\
    Compute $\mathcal{L}(x)$. \\
    \uIf{$\mathcal{L}(x) < \mathcal{L}_{min}$}{
        Return current values of $\theta$, $\tau$, $\phi$. \\
    } 
    \uElse{
        Obtain the adjoint, $\lambda$, by solving the following IVP backwards in time:
        \begin{equation*}
        \begin{aligned}
            \dot{\lambda}(t) & = \nabla_x \ell \left( x(t) \right) - \left[ \partial_x F\left(x(t), x(t - \tau), \tau, t, \theta\right) \right]^T \lambda(t) \\
            &\quad - \mathbb{1}_{t < T - \tau}(t) \left[ \partial_y F \left( x(t + \tau), x(t), \tau, t + \tau, \theta\right) \right]^T \lambda\left(t + \tau \right) \qquad \ t \in [0, T]  \\
            \lambda(T) &= -\nabla_x G \left(x(T)\right)
        \end{aligned}
        \end{equation*} \\
        Compute $\nabla_{\theta} \mathcal{L}(x)$, $\partial \mathcal{L}(x) / \partial \tau$, $\nabla_{\phi} \mathcal{L}(x)$ using equation \eqref{eq:gradients}. \\
        Update $\theta$, $\tau$, and $\phi$ using \texttt{ADAM}. \\
    } 
} 
\end{algorithm}

\begin{table}[hbt]
    \centering 
    \rowcolors{2}{cyan!10}{white}
    
    \begin{tabulary}{1.0\linewidth}{p{1.5cm}L}
        \toprule[0.3ex]
        \textbf{Notation} & \textbf{Meaning} \\
        \midrule[0.1ex]
        
        $\tilde{x}$ & The target trajectory; a cubic-spline interpolation of the noisy data, $\{ \tilde{x}(t_i) \}_{i = 1}^{N_{Data}}$. Ideally, this is a noisy approximation of the true trajectory. \\
        \addlinespace[0.4em]
        $N_{Epochs}$ & The number of epochs we train $\theta$, $\tau$, and $\phi$ for. \\
        \addlinespace[0.4em]
        $\Delta t$ & The time step we use to solve the IVP and generate the predicted trajectory. \\
        \addlinespace[0.4em]
        $N_{\tau}$ & The natural number such that $\tau = N_{\tau} \Delta t$. This is a user selected hyperparameter. \\
        \addlinespace[0.4em]
        $N_{step}$ & The number of steps we take during the forward pass to compute the numerical approximation to the predicted solution. We define $N_{step} = \min\{ n : n \Delta t \geq T \}$. \\
        \addlinespace[0.4em]
        $\mathcal{L}_{min}$ & The loss threshold. This is a user-specified hyper parameter. If the loss drops below this value, we stop training. \\
        \bottomrule[0.3ex]
    \end{tabulary}
    
    \caption{The notation and terminology of section (\ref{Sec:Method})} 
    \label{Table:Method:Notation}
\end{table}
\section{Related Work}
\label{Sec:Background}

In this section, we highlight recent work in learning differential equation parameters using measurements of a solution to that differential equation.
Specifically, we focus on methods that use an adjoint.
There are other approaches, not based on an adjoint, to addressing the problem we consider in this paper (see section \ref{Sec:Problem}).
For instance, the popular \texttt{SINDY} \cite{brunton2016discovering} algorithm can identify sparse, human-readable dynamical systems from noisy measurements of its solution.
\texttt{SINDY} learns differential equations using numeric differentiation and sparse regression.
\texttt{PDE-FIND} \cite{rudy2017data} adapts the \texttt{SINDY} approach to learn PDEs.
Nonetheless, because \texttt{DDE-Find} uses an \emph{optimize-then-discretize} (adjoint) approach, we focus our attention on algorithms that use a similar approach.

\bigskip

While adjoint methods have been popular in the control and numerical differential equation communities for decades (see, for instance, \cite{pontryagin2018mathematical} and \cite{cao2003adjoint}), they were only recently applied to machine learning problems.
One of the landmark attempts to bridge differential equations and machine learning was \cite{weinan2017proposal}.
They noted that residual neural networks \cite{he2016deep} can be interpreted as a forward Euler discretization of continuous time dynamical systems.
They also hypothesized that burgeoning scientific machine learning efforts could exploit this connection to develop specialized architectures.
Shortly after, \cite{chen2018neural} introduced the \texttt{Neural Ordinary Differential Equations} (NODE) framework.
Their approach, built around an adjoint, proposed a method to learn the right-hand side of a differential equation such that given an initial condition, $x_0$, the solution to the IVP defined by the learned equation and specified initial matches a target value at time $T > 0$.
More specifically, they proposed a method to learn a set of parameters, $\theta \in \mathbb{R}^d$, in a parameterized function, $F : \mathbb{R}^d \times \mathbb{R}^p \to \mathbb{R}^d$, such that given an initial condition , $x_0 \in \mathbb{R}^d$, and a target value, $y$, we have $x(T) \approx y$.
Here, $x : [0, T] \to \mathbb{R}^d$ is the solution to the following IVP:
\begin{equation*}
\begin{aligned}
    \dot{x}(t) &= F\left( x(t), \theta \right) \qquad t \in [0, T] \\
    x(0) &= x_0
\end{aligned}
\end{equation*}
Notably, the authors of the NODE framework did not propose it as a tool for learning scientific models from data.
Instead, the authors showed it was a capable machine learning model for discriminative tasks (supervised learning) and generative tasks such as normalizing flows.

\bigskip

Several extensions and follow-ups followed the release of the NODE framework.
One notable example is \cite{rackauckas2020universal}, which emphasizes the potential of NODEs in scientific machine learning.
Specifically, they introduced the \texttt{DifferentialEquations.jl} library, which implemented adjoint approaches for learning differential equations parameters from scientific data.
Another notable example is \cite{ayed2019learning}, which extended the NODE framework to when the user only has partial measurements (some of the components) of the differential equation solution.

\bigskip 

The NODE framework has been extended to other classes of differential equations, including Partial Differential Equations (PDEs) \cite{sun2020neupde, gelbrecht2021neural} and Delay Differential Equations (DDEs) \cite{anumasa2021delay, zhu2021neural}.
The motivation and use cases of both DDE approaches closely mirror those of the original NODE framework.
Namely, they use the learned DDE to map inputs, $x_0 \in \mathbb{R}^d$ (which they use as a constant initial condition in a DDE model), to target values, $y \in \mathbb{R}^d$.
Specifically, they learn a set of parameters, $\theta$, in a parameterized model, $F : \mathbb{R}^d \times \mathbb{R}^d \times \mathbb{R}^p \to \mathbb{R}^d$, such that given an input, $x_0 \in \mathbb{R}^d$, and a target value, $y \in \mathbb{R}^d$, we have $x(T) \approx y$, where $x$ is the solution to the following DDE:
\begin{equation*}
\begin{aligned}
    \dot{x}(t) &= F(x(t), x(t - \tau), \theta) \qquad  &&t \in [0, T] \\
    x(t) &= x_0  \qquad &&t \in [-\tau, 0]
\end{aligned}
\end{equation*}
Both approaches use an adjoint-based approach to learn an appropriate set of parameters, $\theta$.
Notably, both papers use their approaches to solve discriminative tasks, such as image classification.
Further, both models can not learn $\tau$ or the initial condition function.
Nonetheless, they represented an important leap from ODEs to DDEs.

\bigskip 

Another significant advance came with \cite{sandoz2023sindy}, which extended the \texttt{SINDY} framework \cite{brunton2016discovering} to delay differential equations.
Further, unlike previous attempts, their approach can approximately learn the delay from the data.
Their approach does not use an adjoint-based approach; instead, they use a grid search over possible delay values.
Specifically, they assume they have noisy measurements of a function, $x : \mathbb{R} \to \mathbb{R}^d$ at times $t = k \Delta t, k \in \mathbb{Z}$.
They also assume $x$ is a solution to a DDE of the form $\dot{x}(t) = F(x(t), x(t - \tau), \theta)$. 
They assume the delay, $\tau$, is a multiple of $\Delta t$.
They begin with a collection of possible multiples, $s_1, \ldots, s_K$.
For each multiple, $s_k$, they set up a SINDY problem assuming $\tau = s_k \Delta t$.
They assume each component function of $F$ is a linear combination of library terms, where each library term is a monomial of the components of $x(t)$ and $x(t - \tau)$.
They use the data and this assumed model to set up a linear system of equations (where the unknown represents the coefficients in the linear combination for each component of $F$).
They use a greedy sparse regression algorithm to find a sparse solution to this linear system.
The sparse solution gives them an expression for each component function of $F$, effectively learning the dynamics from the data.
They repeat this process for each possible delay and then select the one whose corresponding sparse solution engenders the lowest least squares residual.
Their approach can learn simple DDEs --- including the delay --- from data.
The primary downside to this approach is that if the delay is not a scalar multiple of $\Delta t$, the learned delay may be a bad approximation of the actual delay.
This deficiency can be especially problematic if the gradient of the loss function is large near the true parameter values (something we empirically found to be true in the experiments we considered in section \ref{Sec:Experiments}).
In this case, the grid search may not be fine enough to catch the steep minimum, possibly leading the algorithm to suggest an inaccurate delay.
Nonetheless, the significance of learning the delay --- even if only approximately --- from data can not be understated.

\bigskip 

Finally, \cite{stephany2024learning} recently proposed an algorithm to learn the parameters, $\theta$, and delay, $\tau$ in a DDE from data.
Specifically, they assume they have measurements of a function $x$ which solves a DDE of the form
\begin{equation*}
\begin{aligned}
    \dot{x}(t) &= F(x(t), x(t - \tau), \theta) \qquad && t \in [0, T] \\
    x(t) &= x_0 && t \in [-\tau, 0].
\end{aligned}
\end{equation*}
The paper proposes a novel adjoint method to learn $\theta$ and $\tau$ from measurements of $x$.
They demonstrated the method's ability to learn simple DDE models, including the delay.
They also rigorously justify their adjoint approach.
However, their approach is limited to constant initial conditions, can not learn the initial conditions, and can not handle the case when $F$ directly depends on $\tau$.
Further, their implementation can only deal with the case $d = 1$, and the paper did not explore the proposed algorithm's robustness to noise.
Nonetheless, \cite{stephany2024learning} is the first approach that can learn $\tau$ without restricting the possible values of $\tau$.

\bigskip

\texttt{DDE-Find} builds upon this work by weakening the assumptions on the underlying DDE ($F$ can directly depend on $\tau$, and $X_0$ can be an arbitrary, unknown function).
Thus, we extend the work of \cite{stephany2024learning} for learning $\tau$ to a general-purpose algorithm that can learn $\theta$, $\tau$, and $\phi$ simultaneously from noisy measurements of $x$.
Further, our implementation allows $F$ and $X_0$ to be any \texttt{torch.nn.Module} object (effectively any function that you can implement in \texttt{Pytorch}) that satisfies assumptions \ref{assumption:FGell} and \ref{assumption:X0}. 
\section{Experiments}
\label{Sec:Experiments}

In this section, we test \texttt{DDE-Find} on a collection of DDEs.
These experiments showcase \texttt{DDE-Find}'s ability to learn $\theta$, $\tau$, and $\phi$ directly from noisy measurements of the DDE's solution.
Our implementation, which is available at \url{https://github.com/punkduckable/DDE-Find}, is built on top of \texttt{PyTorch} \cite{paszke2019pytorch}.
\texttt{DDE-Find} implements $F$ and $X_0$ as trainable \texttt{torch.nn.Module} objects.
Thus, the user can use any \texttt{torch.nn.Module} objects that satisfy assumptions \ref{assumption:FGell} and \ref{assumption:X0}, respectively, for $F$ and $X_0$.
In particular, $F$ and $X_0$ can be structured models with fixed parameters, deep neural networks, or anything in between.

\bigskip 

In these experiments, we focus on the case where $F$ and $X_0$ are structured parameter models.
We demonstrate \texttt{DDE-Find}'s ability to learn the parameters in these models.
By a \emph{structured model}, we mean an expression of the form 
\begin{equation}
\label{eq:F:Structured}
    F(x(t), x(t - \tau), \tau, t, \theta) = \frac{\sum_{i = 1}^{n_a} a_i f_i\left(x(t), x(t - \tau), \tau, t\right)}{\sum_{j = 1}^{n_b} b_j g_j\left(x(t), x(t - \tau), \tau, t\right) } + \sum_{k = 1}^{n_c} c_k h_k\left(x(t), x(t - \tau), \tau, t \right).
\end{equation}
In this case,
$$\theta = \text{concat}\left[ \left(a_1, \ldots, a_{n_a} \right), \left(b_1, \ldots, d_{n_b} \right), \left( c_1, \ldots, c_{n_c} \right) \right] \in \mathbb{R}^{n_a + n_b + n_c},$$
and the functions $h_i$, $g_j$, and $h_k$ are user-selected.
Thus, for structured prediction models, learning $\theta$ involves finding a set of $a_i, b_j$, and $c_k$ such that if we substitute equation \eqref{eq:F:Structured} into the IVP, equation \eqref{eq:IVP}, and then solve, the resulting solution closely matches the noisy measurements.

\bigskip 

We also use structured models for $X_0$.
Specifically, we consider affine and periodic models for $X_0$.
In the affine model, $X_0(t) = at + b$, with $a, b \in \mathbb{R}^{d}$.
In this case $\phi = \text{concat}[a, b] \in \mathbb{R}^{2d}$.
For the sinusoidal model, $X_0(t) = A \odot \sin(\omega t) + b$, with $A, \omega, b \in \mathbb{R}^{d}$ (we apply $\sin$ element-wise to $\omega t \in \mathbb{R}^d$. Also, $\odot : \mathbb{R}^d \times \mathbb{R}^d \to \mathbb{R}^d$ denotes element-wise multiplication).
In this case, $\phi = \text{concat}[A, \omega, b] \in \mathbb{R}^{3d}$.
In addition to these models, our implementation allows $X_0$ to be a constant function or an arbitrary neural network. \texttt{DDE-Find} supports using constant functions and an arbitrary neural network for $X_0$.

\bigskip 

We use structured models (as opposed to arbitrary deep neural networks) in our experiments because it is easier to elucidate how \texttt{DDE-Find} works using these models.
Further, using neural networks for $F$ or $X_0$ can over-parameterize the model, engendering non-uniqueness issues.
We explore this issue in detail in section \ref{Sec:Discussion:Neural}.
Moreover, because \texttt{DDE-Find} uses $F$ to solve the IVP, equation \eqref{eq:IVP}, an improperly initialized $F$ network can engender an IVP whose solution grows exponentially and exceeds the limit of the 32-bit floating point range before $t = T$.
Thus, using a neural network for $F$ or $X_0$ requires careful initialization strategies that are beyond the scope of this paper.
We discuss this challenge and potential future work to address it in section \ref{Sec:Discussion:Neural}.
Notwithstanding, \texttt{DDE-Find} supports using neural networks for $F$ and $X_0$, should the reader want to use them.

\bigskip 

In our experiments, we select the noise to be proportional to the sample standard deviation of the \emph{noise-free dataset}, $\{ x(t_i) \}_{i = 1}^{N_{Data}}$ (where $x$ denotes the true trajectory --- the solution to the IVP, equation \eqref{eq:IVP}, that generated the noisy dataset).
Specifically, we say a noisy data set has a \emph{noise level} of $l > 0$ if $\sigma = l * \sigma_{Data}$, where $\sigma_{Data}$ is the (unbiased) sample standard deviation of the noise-free dataset and $\sigma$ is the standard deviation of the noise random variables (see section \ref{Sec:Problem}).

\bigskip 

We use the \texttt{Adam} optimizer \cite{kingma2014adam} throughout our experiments, though our implementation supports other optimizers (\emph{e.g.} \texttt{RMSProp}, \texttt{LBFGS}, \texttt{SGD}).
The \texttt{Adam} optimizer has two $\beta$ hyperparameters. 
For these experiments, we use the default parameter values in \texttt{PyTorch} ($\beta_1 = 0.9, \beta_2 = 0.999$).
Further, unless otherwise stated, we use the $L^2$ norm for $\ell$ and the zero function for $G$.
Notably, our implementation of \texttt{DDE-Find} supports using (weighted) $L^1$, $L^2$, and $L^{\infty}$ norms for $\ell$ and $G$ \footnote{Note that the $L^1$ and $L^{\infty}$ norms are not $C^1$, meaning that if we use them, then $\ell$ and $G$ violate assumption \ref{assumption:FGell}. 
However, these norms are continuously differentiable almost everywhere. 
In practice, this discrepancy causes no issues, even if it means we technically can't apply theorem \ref{Theorem:1}.}.
Further, in all experiments, we train for $500$ epochs ($N_{Epochs} = 500$) with a learning rate of $0.03$.
We use a cosine annealing learning rate scheduler. 
We initialized this scheduler such that the final learning rate is $10\%$ of the initial one.
Finally, we use $\mathcal{L}_{min} = 0.01$.

\subsection{Delay Exponential Decay Model}
\label{Sec:Experiments:Exponential}

We begin with the Delay Exponential Decay model, arguably the simplest non-trivial DDE.
This model can model cell growth.
In this case, the delay term models the time required for the cells to mature \cite{rihan2021ddes}.
The IVP for this DDE is
\begin{equation}
\label{eq:IVP:Exponential}
\begin{aligned}
    \dot{x}(t) &= \theta_0 \odot x(t) + \theta_1 \odot x(t - \tau) && t \in [0, T] \\
    x(t) &= X_0(t, \phi) && t \in [-\tau, t].
\end{aligned}
\end{equation}
To generate the true trajectory, we set $d = 1$, $T = 10$, $\tau = 1.0$, and $\theta_0 = \theta_1 = -2.0$.
We also use an affine function for the initial condition. 
Specifically, 
\begin{equation}
\label{Eq:X0:Affine}
    X_0(t, (a, b)) = at + b.
\end{equation}
For our experiments, we set $a = 1.5$ and $b = 4.0$.
We solve this IVP using a second order Runge-Kutta method with a step size of $0.1$. 
The target trajectory contains $N_{Data} = 100$ data points.

\bigskip 

For these experiments, we use structured models derived from equations \eqref{eq:IVP:Exponential} and \eqref{Eq:X0:Affine} for $F$ and $X_0$, respectively.
Thus, for these experiments $\theta = \text{concat}\left[ \theta_0, \theta_1 \right] \in \mathbb{R}^{2d}$ and $\phi = \text{concat}\left[a, b\right] \in \mathbb{R}^{2d}$.

\bigskip 

We run three sets of experiments on this model corresponding to three different noise levels.
Specifically, we consider noise levels of $0.1$, $0.3$, and $0.9$.
This set of noise levels us to analyze \texttt{DDE-Find}'s performance across three orders of magnitude of noise.
For each noise level, we generate $20$ target trajectories by adding Gaussian noise with the specified noise level to the true trajectory.
For each target trajectory, we use \texttt{DDE-Find} to learn $\theta = (\theta_0, \theta_1)$, $\tau$, and $\phi = (a, b)$.
For each experiment, we initialize $\theta$, $\tau$ and $\phi$ to $(-1.5, -2.5)$, $\tau = 2.0$, and $\phi = (2.25, 2.8)$. 
Finally, we use $N_{\tau} = 10$.

\bigskip 

\begin{figure}[hbt!]
    \centering
    \includegraphics{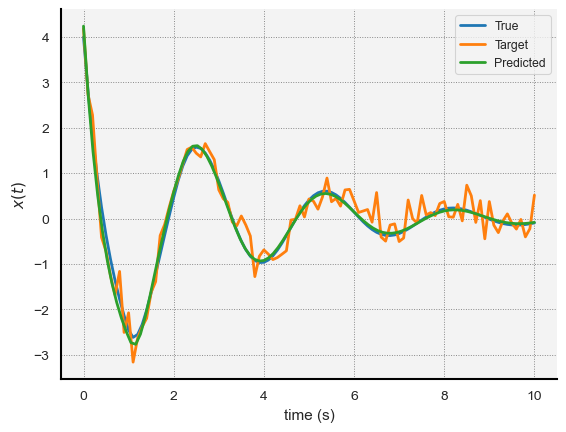}
    \caption{True, Target, and Predicted trajectories from one of the Delay Exponential Decay experiments with a noise level of $0.3$.}
    \label{fig:Exponential}
\end{figure}

Table \ref{table:Exponential:0.3} report the results for the experiments with a noise level of $0.3$.
In appendix \ref{Sec:Tables}, tables \ref{table:Exponential:0.1} and \ref{table:Exponential:0.9} report the results for the experiments with a noise level of $0.1$ and $0.9$, respectively.
Further, figure \ref{fig:Exponential} shows the true, target, and predicted trajectories for one of the experiments with a noise level of $0.3$.

\bigskip

\begin{table}[hbt!]
    \centering
    \caption{Results for the Delay Exponential Decay Equation for noise level $0.3$}
    \rowcolors{2}{white}{olive!20!green!15}

    \begin{tabulary}{\linewidth}{C C C C}
        \toprule[0.3ex]
        Parameter &  True Value & Mean Learned Value & Standard Deviation \\
        \midrule[0.1ex]
        $\theta_0$  & $-2.0$    & $-2.0454$ & $0.0547$ \\
        $\theta_1$  & $-2.0$    & $-2.0243$ & $0.0397$ \\
        $\tau$      & $1.0$     & $1.0104$  & $0.0117$ \\
        $a$         & $1.5$     & $1.4339$  & $0.2261$ \\
        $b$         & $4.0$     & $4.0001$  & $0.1350$ \\
        \bottomrule[0.3ex]    
    \end{tabulary}
        
    \label{table:Exponential:0.3}
\end{table}

Across all experiments, \texttt{DDE-Find} can consistently learn $\theta$, $\tau$, and $\phi$.
This holds even when we corrupt the target trajectory corrupted with high levels of noise.
Unsurprisingly, as the noise level increases, the accuracy of the learned parameters tends to decrease.
However, even with a noise level of $0.9$ (meaning the standard deviation of the noise is nearly as great as that of the underlying dataset), the relative error between the true and learned parameters is consistently only a few percent. 
\texttt{DDE-Find} is especially good at learning $\tau$; even at a noise level of $0.9$, the mean learned $\tau$ value barely differs from the true value.
In almost every experiment, the learned $\tau$ value is within $5\%$ of the true value of $\tau$, despite being initialized to double the true value.
\texttt{DDE-Find} has more trouble recovering $a$ in the high-noise experiments.
This isn't surprising, however, as empirical tests reveal that the solution to equation \eqref{eq:IVP:Exponential} is insensitive to changes in $a$.
We can also explain this result theoretically by considering the form of the gradient for $\phi$ (in equation \eqref{eq:gradients}) and the affine ICs.
We discuss this in greater detail in section \ref{Sec:Discussion:phi}.
Notwithstanding, \texttt{DDE-Find} performs well in all experiments with the Delay Exponential Decay equation.

\subsection{Logistic Delay Model}
\label{Sec:Experiments:Logistic}

Next, we consider the Logistic Delay Equation.
This non-linear DDE can model various oscillatory phenomenon in ecology that limit to a stable equilibrium population \cite{murray2003mathematical} \cite{baker2020global}.
For this equation, $d = 1$.
The IVP for this DDE is
\begin{equation}
\label{eq:IVP:Logistic}
\begin{aligned}
    \dot{x}(t) &= \theta_0  x(t) \left( 1 - \theta_1 x(t - \tau) \right) && t \in [0, T] \\
    x(t) &= X_0(t, \phi) && t \in [-\tau, t].
\end{aligned}
\end{equation}
To generate the true trajectory, we set $T = 10$, $\tau = 1.0$, and $\theta_0 = 2.0$, and $\theta_1 = 1.5$.
For these experiments, we use a periodic initial condition:
\begin{equation}
\label{Eq:X0:Periodic}
    X_0(t, (a, b)) = A \odot \sin(\omega t ) + b
\end{equation}
We set $a = -0.5$, $\omega = 5.0$, and $b = 2.0$.
We solve this IVP using a second order Runge-Kutta method with a step size of $0.1$. 
The target trajectory contains $N_{Data} = 100$ data points.
For these experiments, we use structured models derived from equations \eqref{eq:IVP:Logistic} and \eqref{Eq:X0:Periodic} for $F$ and $X_0$, respectively.

\bigskip 

As with the Delay Exponential Decay equation, we run experiments at 3 different noise levels, $0.1$, $0.3$, and $0.9$.
For each noise level, we generate $20$ target trajectories by adding Gaussian noise with the specified noise level to the true trajectory.
For each target trajectory, we use \texttt{DDE-Find} to learn $\theta = (\theta_0, \theta_1)$, $\tau$, and $\phi = (A, \omega, b)$.
For each experiment, we initialize $\theta = (1.0, 1.0)$, $\tau = 0.5$, and $\phi = (-0.25, 6.0, 2.6)$.
Finally, we use $N_{\tau} = 10$.

\bigskip 

\begin{figure}[hbt!]
    \centering
    \includegraphics{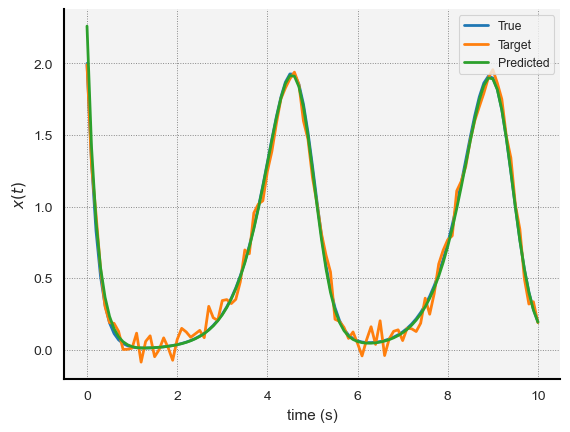}
    \caption{True, Target, and Predicted trajectories from one of the Logistic Delay experiments with a noise level of $0.3$.}
    \label{fig:Logistic}
\end{figure}

Table \ref{table:Logistic:0.3} report the results for the experiments with a noise level of $0.3$.
In appendix \ref{Sec:Tables}, tables \ref{table:Logistic:0.1} and \ref{table:Logistic:0.9} report the results for the experiments with a noise level of $0.1$ and $0.9$, respectively.
Further, figure \ref{fig:Logistic} shows the true, target, and predicted trajectories for one of the experiments with a noise level of $0.3$.

\bigskip

\begin{table}[hbt!]
    \centering
    \caption{Results for the Logistic Decay Equation for noise level $0.3$}
    \rowcolors{2}{olive!20!green!15}{white}

    \begin{tabulary}{\linewidth}{C C C C}
        \toprule[0.3ex]
        Parameter &  True Value & Mean Learned Value & Standard Deviation \\
        \midrule[0.1ex]
        $\theta_0$  & $2.0$     & $1.9928$ & $0.0740$ \\
        $\theta_1$  & $1.5$     & $1.5145$ & $0.0416$ \\
        $\tau$      & $1.0$     & $0.9998$ & $0.0134$ \\
        $A$         & $-0.5$    & $0.2911$ & $0.5485$ \\
        $\omega$    & $3.0$     & $5.8176$ & $0.4387$ \\
        $b$         & $2.0$     & $2.3938$ & $0.2292$ \\
        \bottomrule[0.3ex]    
    \end{tabulary}
        
    \label{table:Logistic:0.3}
\end{table}

Once again, \texttt{DDE-Find} successfully learns a DDE whose predicted trajectory almost perfectly matches the true one, notwithstanding noise in the data.
Further, across all experiments, the learned $\theta$ and $\tau$ values closely match the true values.
With that said, \texttt{DDE-Find} has some trouble recovering the parameters in the initial condition.
While \texttt{DDE-Find} consistently is able to identify the offset parameter $b$, it has trouble identifying $A$ and $\omega$. 
As before, we believe this is because the trajectory is very insensitive to these parameters.
Further, in section \ref{Sec:Discussion:phi}, we provide a theoretical explanation for this phenomenon using the expression for $\nabla_{\phi} \mathcal{L}$ in equation \eqref{eq:gradients}.
Nonetheless, the predicted trajectory in \ref{fig:Logistic} almost perfectly matches the true trajectory and the learned $\tau$, $\theta$ values are spot-on in almost every experiment.

\subsection{El Ni\~no / Southern Oscillation Model}
\label{Sec:Experiments:ENSO}

We now consider a model for the El Ni\~no / Southern Oscillation (ESNO).
El Ni\~no is a periodic variation in oceanic current and wind patterns in the Pacific Ocean.
Here, we consider a model for sea surface temperature anomaly arising from El Ni\~no.
Specifically, we consider a modified version of a model originally proposed in \cite{suarez1988delayed}.
For this DDE, $d = 1$.
The IVP for this DDE is
\begin{equation}
\label{eq:IVP:ENSO}
\begin{aligned}
    \dot{x}(t) &= \theta_0 x(t) - \theta_1 x(t)^3 - \theta_2 x(t - \tau)&& t \in [0, T] \\
    x(t) &= X_0(t, \phi) && t \in [-\tau, t].
\end{aligned}
\end{equation}
To generate the true trajectory, we set $T = 10$, $\tau = 5.0$, and $\theta_0 = 1.0$, $\theta_1 = 1.0$, and $\theta_2 = 1.5$.
As in the Logistic Delay equation experiments, we use periodic initial conditions.
For these experiments, we set $a = -0.25$, $\omega = 1.0$, and $b = 1.5$.
We solve this IVP using a second order Runge-Kutta method with a step size of $0.1$. 
The target trajectory contains $N_{Data} = 100$ data points.
For these experiments, we use structured models derived from equations \eqref{eq:IVP:ENSO} and \eqref{Eq:X0:Periodic} for $F$ and $X_0$, respectively.

\bigskip 

We run experiments at 3 different noise levels, $0.1$, $0.3$, and $0.9$.
For each noise level, we generate $20$ target trajectories by adding Gaussian noise with the specified noise level to the true trajectory.
For each target trajectory, we use \texttt{DDE-Find} to learn $\theta = (\theta_0, \theta_1, \theta_2)$, $\tau$, and $\phi = (A, \omega, b)$.
For each experiment, we initialize $\theta = (1.5, 0.8, 1.2)$, $\tau = 6.0$, and $\phi = (-0.3, 0.8, 1.95)$.
Finally, we use $N_{\tau} = 50$.

\bigskip 

\begin{figure}[hbt!]
    \centering
    \includegraphics{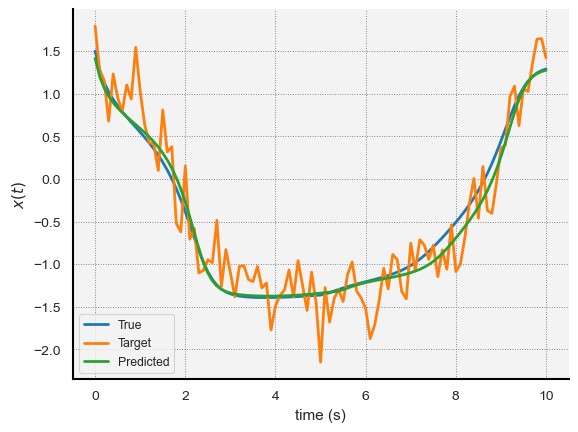}
    \caption{True, Target, and Predicted trajectories from one of the ENSO experiments with a noise level of $0.3$.}
    \label{fig:ENSO}
\end{figure}

Table \ref{table:ENSO:0.3} report the results for the experiments with a noise level of $0.3$.
In appendix \ref{Sec:Tables}, tables \ref{table:ENSO:0.1} and \ref{table:ENSO:0.9} report the results for the experiments with a noise level of $0.1$ and $0.9$, respectively.
Further, figure \ref{fig:ENSO} shows the true, target, and predicted trajectories for one of the experiments with a noise level of $0.3$.

\bigskip

\begin{table}[hbt!]
    \centering
    \caption{Results for the ENSO for noise level $0.3$}
    \rowcolors{2}{white}{olive!20!green!15}

    \begin{tabulary}{\linewidth}{C C C C}
        \toprule[0.3ex]
        Parameter &  True Value & Mean Learned Value & Standard Deviation \\
        \midrule[0.1ex]
        $\theta_0$  & $1.0$     & $1.1038$. & $0.1745$ \\
        $\theta_1$  & $1.0$     & $1.2172$  & $0.1561$ \\
        $\theta_2$  & $0.75$    & $0.8174$  & $0.0810$ \\
        $\tau$      & $5.0$     & $5.2157$  & $0.2512$ \\
        $A$         & $-0.25$   & $-0.6633$ & $0.1844$ \\
        $\omega$    & $1.0$     & $0.9031$  & $0.1430$ \\
        $b$         & $1.5$     & $1.4933$  & $0.1788$ \\
        \bottomrule[0.3ex]    
    \end{tabulary}
        
    \label{table:ENSO:0.3}
\end{table}

\texttt{DDE-Find} consistently identifies the correct $\tau$ and $\theta$ values, even when faced with extreme noise.
\texttt{DDE-Find} also reliably recovers $b$ the initial condition, but has trouble recovering$A$.
Empirical tests reveal that the true trajectory is insensitive to $A$.
We justify this result theoretically in section \ref{Sec:Discussion:phi}.
Nonetheless, across all experiments, the predicted trajectory from the final model almost perfectly matches the target trajectory.

\subsection{Cheyne–Stokes Respiration Model}
\label{Sec:Experiments:Chyene}

Cheyne-Stokes respiration is a human aliment characterized by an irregular breathing pattern \cite{murray2003mathematical}.
People suffering from this condition exhibit progressively deeper and faster breaths, followed by a gradual decrease.
The blood CO2 levels of people suffering from Cheyne stokes can be modeled by the following delay differential equation \cite{murray2003mathematical}:
\begin{equation}
\label{eq:IVP:Cheyne}
\begin{aligned}
    \dot{x}(t) &= p - V_0 x(t)\frac{x^m(t - \tau)}{\alpha + x^m(t - \tau)} & \qquad t \in [0, T] \\
    x(t) &= X_0(t, \phi) & \qquad t \in [-\tau, 0].
\end{aligned}
\end{equation}
For this equation, $d = 1$ and $\theta = (p, V_0, a)$.
We treat $m \in \mathbb{N}$ as a fixed constant, not a trainable parameter.
For our experiments, we set $m = 8$.

\bigskip 

To generate the true trajectory, we set $T = 3.0$, $\tau = 0.25$, $p = 1.0$, $V_0 = 7.0$, and $\alpha = 2.0$.
As in the Delay Exponential Decay experiments, we use affine initial conditions.
Specifically, we set $a = -5.0$ and $b = 2.0$ (in $X_0(t, \phi) = at + b$).
Using these parameters, we solve equation \eqref{eq:IVP:Cheyne} using a second order Runge-Kutta method with a step size of $0.025$.
Thus, for these experiments, the target trajectory contains $N_{Data} = 120$ data points.
We use structured models derived from equations \eqref{eq:IVP:Cheyne} and \eqref{Eq:X0:Affine} for $F$ and $X_0$, respectively.

\bigskip 

We run experiments at 3 different noise levels, $0.1$, $0.3$, and $0.9$.
For each noise level, we generate $20$ target trajectories by adding Gaussian noise with the specified noise level to the true trajectory.
For each target trajectory, we use \texttt{DDE-Find} to learn $\theta = (p, V_0, \alpha)$, $\tau$, and $\phi = (a, b)$. 
For each experiment, we initialize $\theta = (2.0, 12.0, 1.5)$ and $\tau = 0.5$. 
Finally, to initialize $\phi$, we set $a = 0$ and set $b$ to the first noisy data point, $\tilde{x}(t_1)$ (this means that we use a different initial $\phi$ guess each experiment).
Finally, we use $N_{\tau} = 10$. 

\bigskip 

\begin{figure}[hbt!]
    \centering
    \includegraphics{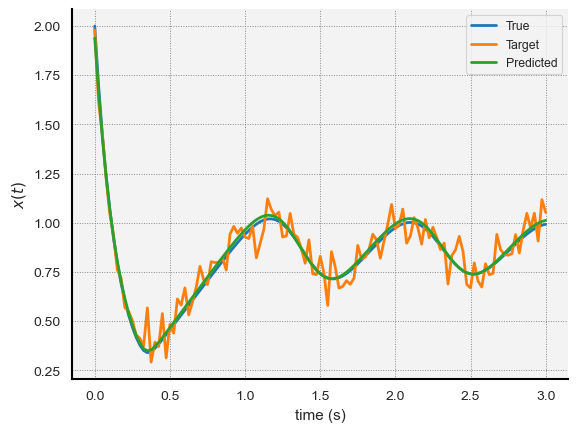}
    \caption{True, Target, and Predicted trajectories from one of the Cheyne experiments with a noise level of $0.3$.}
    \label{fig:Cheyne}
\end{figure}

Table \ref{table:Cheyne:0.3} report the results for the experiments with a noise level of $0.3$.
In appendix \ref{Sec:Tables}, tables \ref{table:Cheyne:0.1} and \ref{table:Cheyne:0.9} report the results for the experiments with a noise level of $0.1$ and $0.9$, respectively.
Further, figure \ref{fig:Cheyne} shows the true, target, and predicted trajectories for one of the experiments with a noise level of $0.3$.

\bigskip

\begin{table}[hbt!]
    \centering
    \caption{Results for the Cheyne-Stoke Respiration model for noise level $0.3$}
    \rowcolors{2}{olive!20!green!15}{white}

    \begin{tabulary}{\linewidth}{C C C C}
        \toprule[0.3ex]
        Parameter &  True Value & Mean Learned Value & Standard Deviation \\
        \midrule[0.1ex]
        $p$         & $1.0$     & $1.0114$  & $0.0229$ \\
        $V_0$       & $7.0$     & $7.5973$  & $0.2567$ \\
        $\alpha$    & $2.0$     & $2.2845$  & $0.3004$ \\
        $\tau$      & $0.25$    & $0.2474$  & $0.0037$ \\
        $a$         & $-5.0$    & $2.4795$  & $0.3232$ \\
        $b$         & $2.0$     & $1.9236$  & $0.1048$ \\
        \bottomrule[0.3ex]    
    \end{tabulary}
        
    \label{table:Cheyne:0.3}
\end{table}

Thus, \texttt{DDE-Find} can consistently identify the correct $\tau$, and $\theta$ values across a wide range of noise levels.
It can also reliably identify the offset ($b$) of the initial condition, though it struggles to identify the correct $a$ value.
Further, at high noise levels, the relative error in the mean learned $\alpha$ and $V_0$ values are somewhat high.
Nonetheless, the identified parameters are still fairly close to their true values, despite being initialized with relatively poor estimates for each parameter.
As with the other experiments, empirical tests reveals that there is a correlation between how sensitive the true trajectory is to perturbations in a particular parameter and the relative error in the mean value that \texttt{DDE-Find} learns for that parameter.
We explore this phenomenon in greater depth in section \ref{Sec:Discussion:phi}. 
Nonetheless, across all experiments, the predicted trajectory from the final model almost perfectly matches the target trajectory.

\subsection{HIV Model}
\label{Sec:Experiments:HIV}

Finally, we turn our attention to a delay model for Human Immunodeficiency Viruses (HIV).
This model was originally proposed in \cite{nelson2000model} to describe the number of HIV-infected T-cells over time when a drug partially inhibits the virus.
Unlike the previous models we have considered, this one takes place in $\mathbb{R}^3$ ($d = 3$) and $F$ features an explicit dependence on $\tau$.
The IVP for this model is
\begin{equation}
\label{eq:IVP:HIV}
\begin{aligned}
    \dot{T^*}(t) &= k T_0 V_{I}(t - \tau) \exp(-m \tau) - \delta T^{*}(t) \\
    \dot{V_{I}}(t) &= (1 - n_p) N \delta T^*(t) - c V_I(t) &\qquad t \in [0, T] \\
    \dot{V_{NI}}(t) &= n_p N \delta T^*(t) - c V_{NI}(t) \\
    \begin{bmatrix} T^*(t) \\ V_{I}(t) \\ V_{NI}(t) \end{bmatrix} &= X_0(t, \phi) & \qquad t \in [-\tau, 0].
\end{aligned}
\end{equation}
Here, $T_0$ denotes the number of non-infected T-cells (which we assume is constant), $T^*$ denotes the number of infected T-cells, $V_{I}$ denotes the number of infectious viruses, and $V_{NI}$ denotes the number of non-infectious viruses.
Notice that the $T^*$ equation explicitly depends on $\tau$.
\cite{nelson2000model} showed that this DDE can model the dynamics of HIV infections.
Unfortunately, the model is over-parameterized.
Specifically, given some $\eta \in \mathbb{R} - \{0\}$, if we replace $kT_0$ with $kT_0 \exp(-\tau \eta)$ and $m$ with $m - \eta$, respectively, the model remains identical.
Further, replacing $N$ and $\delta$ with $N/\eta$ and $\delta \eta$, then the second and third equations remain unchanged.
Empirical tests also reveal that given a particular target trajectory, there are many distinct parameter combinations that engender (approximately) the same trajectory.
We could remedy this by fixing some of the parameters and learning the rest.
Instead, for these experiments, we treat the model's parameters as known constants and use \texttt{DDE-Find} to learn $\tau$ and $\phi$.

\bigskip 

To generate the true trajectory, we set $T = 10.0$, $\tau = 1.0$, $k = 0.00343$, $m = 3.8$, $d = 0.05$, $c = 2.0$, $T_0 = 1000.0$, $n_p = 0.43$, and $N = 48.0$. 
These values are loosely based on values in \cite{nelson2000model}.
As in the Delay Exponential Decay experiments, we use periodic initial conditions.
Specifically, we set $A = (1.0, 3.0, 0.0)$, $\omega = (1.0, 3.0, 0.0)$, and $b = (10.0, 8.0, 0.0)$ (in $X_0(t, \phi) = A \odot \sin(\omega t) + b$).
Using these parameters, we solve equation \eqref{eq:IVP:HIV} using a second order Runge-Kutta method with a step size of $0.1$.
Thus, the target trajectory contains $N_{Data} = 100$ data points.
For these experiments, we use structured models derived from equations \eqref{eq:IVP:HIV} and \eqref{Eq:X0:Periodic} for $F$ and $X_0$, respectively.

\bigskip 

We run experiments at 3 different noise levels: $0.0$, $0.3$, and $0.9$.
For these experiments, we use the $L^1$ norm for $\ell$.
Note that this means $\ell$ does not satisfy assumption \ref{assumption:FGell} (since the map $x \to \| x \|_1$ is not continuously differentiable along the coordinate axes).
We demonstrate that \texttt{DDE-Find} nonetheless identifies the correct $\tau$ and $\phi$ values.
For each noise level, we generate $20$ target trajectories by adding Gaussian noise with the specified noise level to the true trajectory.
For each target trajectory, we use \texttt{DDE-Find} to learn $\tau$, and $\phi = (A, \omega, b) \in \mathbb{R}^9$. 
For each experiment, we initialize $\tau = 0.25$, $A = (1, 1, 1)$, $\omega = (1.2, 3.6, 0)$, and $b$ to the first noisy data point, $\tilde{x}(t_1) \in \mathbb{R}^3$ (this means that we use a different initial $\phi$ guess each experiment).
Finally, we use $N_{\tau} = 10$. 

\bigskip 

\begin{figure}[H]
    \centering
    \includegraphics[width=0.73\textwidth]{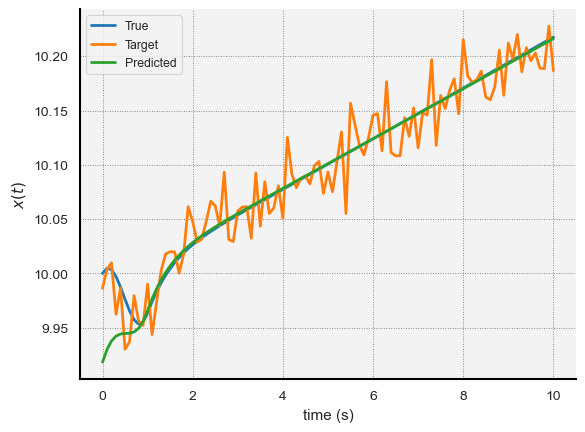}
    \caption{The $T^*$ component of the true, target, and predicted trajectory for the HIV model.}
    \label{fig:HIV:T*}
\end{figure}

\begin{figure}[H]
    \centering
    \includegraphics[width=0.73\textwidth]{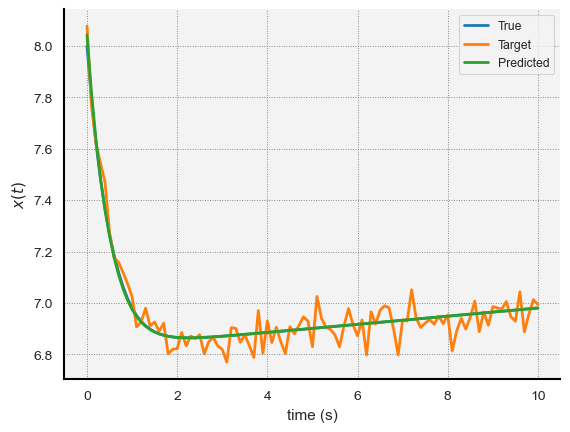}
    \caption{The $V_{I}$ component of the true, target, and predicted trajectory for the HIV model.}
    \label{fig:HIV:V_I}
\end{figure}

\begin{figure}[H]
    \centering
    \includegraphics[width=0.73\textwidth]{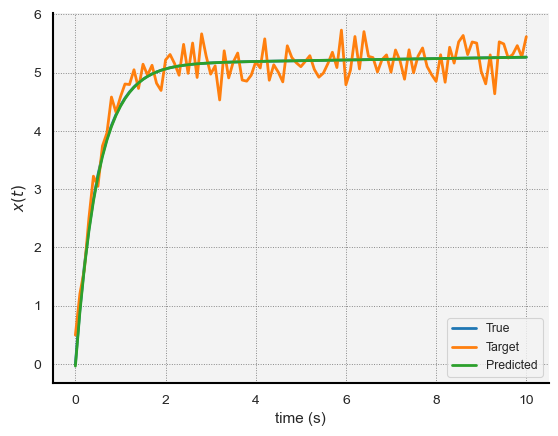}
    \caption{The $V_{NI}$ component of the true, target, and predicted trajectory for the HIV model.}
    \label{fig:HIV:V_NI}
\end{figure}

Table \ref{table:HIV:tau} report the results for $\tau$ across the three experiments.
In appendix \ref{Sec:Tables}, tables \ref{table:HIV:Phi:0.1}, \ref{table:HIV:Phi:0.3}, and \ref{table:HIV:Phi:0.9} report the results for $\phi$ in the experiments with a noise level of $0.1$, $0.3$, and $0.9$, respectively.
The learned $\phi$ values exhibit some very interesting properties that highlight an interesting feature (and limitation) of system identification of DDEs.
We discuss this in section \ref{Sec:Discussion:phi}.
Further, figures \ref{fig:HIV:T*}, \ref{fig:HIV:V_I}, and \ref{fig:HIV:V_NI} show the $T^*$, $V_{I}$, and $V_{NI}$ components of the true, target, and predicted trajectories, respectively, for one of the experiments with a noise level of $0.3$.

\bigskip

\begin{table}[hbt!]
    \centering
    \caption{$\tau$ results for the HIV model}
    \rowcolors{2}{olive!20!green!15}{white}

    \begin{tabulary}{\linewidth}{C C C}
        \toprule[0.3ex]
        Noise Level & Mean Learned Value & Standard Deviation \\
        \midrule[0.1ex]
        True        & $1.0$     & N/A \\
        $0.1$       & $1.0015$     & $0.0008$ \\
        $0.3$       & $1.0011$     & $0.0024$ \\
        $0.9$       & $0.9961$     & $0.0054$ \\
        \bottomrule[0.3ex]    
    \end{tabulary}
        
    \label{table:HIV:tau}
\end{table}

Thus, even when operating in $\mathbb{R}^3$ and when $F$ has an explicit dependence on $\tau$, \texttt{DDE-Find} successfully learns values for $\tau$ and $\phi$ such that the corresponding predicted trajectory closely matches the true one.
Interestingly, while the predicted and true trajectories in figures \ref{fig:HIV:V_I} and \ref{fig:HIV:V_NI} almost perfectly overlap, they noticeably disagree for $t \in [0, 1]$ in figure \ref{fig:HIV:T*}.
We believe this is simply a result of incomplete training; since the $T^*$ trajectory varies the least (only going from $10.0$ at $t = 0$ to about $10.2$ at $t = 10$), a slight mismatch between the predicted and true $T^*$ trajectory only incurs a small increase in the loss;
A proportional mismatch between the other components of the predicted and true trajectories would incur a much larger loss.
Thus, it makes sense that \texttt{DDE-Find} does the worst job recovering the $T^*$ component of the solution --- it has a weak incentive to do so.
Nonetheless, the predicted trajectory still closely matches the true one.

\bigskip 

Across all experiments, the learned $\tau$ value almost perfectly matches the target one, despite being initialized to just $25\%$ of the correct value.
Further, the tables in appendix \ref{Sec:Tables} reveal that \texttt{DDE-Find} consistently identifies $b$ (the offset) in the initial condition function. 
However, \texttt{DDE-Find} is unable to learn the $T^*$ and $V_{NI}$ components of $\omega$ and $A$.
We theoretically justify this surprising result in section \ref{Sec:Discussion:phi}.
Specifically, we show that the gradient of the loss with respect to these components of $\phi$ is identically zero.
Briefly, this phenomenon arises because $A \sin(\omega 0) = 0$, and the HIV model only depends on the $V_{I}$ component of the delayed state.
Crucially, however, \texttt{DDE-Find} learns the delay, $\tau$, and initial condition offset, $b$, across all experiments.

\section{Discussion}
\label{Sec:Discussion}

In this section, we discuss further aspects of the \texttt{DDE-Find} algorithm.
First, in section \ref{Sec:Discussion:Adjoint}, we provide insight into why we use an adjoint to compute $\mathcal{L}$'s gradients.
Next, in section \ref{Sec:Discussion:Ntau}, we discuss the $N_{\tau}$ hyperparameter.
In section \ref{Sec:Discussion:Loss}, we detail how our implementation of \texttt{DDE-Find} evaluates the loss function in practice and elucidate why we chose this approach.
Next, in section \ref{Sec:Discussion:phi}, we analyze why \texttt{DDE-Find} struggles to learn some components of $\phi$.
Finally, in section \ref{Sec:Discussion:Neural}, we discuss how \texttt{DDE-Find} performs when we use a neural network to parameterize $F$ or $X_0$.

\subsection{Do we Really Need an Adjoint?}
\label{Sec:Discussion:Adjoint}

Throughout this paper, we use an adjoint approach to compute the gradients of the loss with respect to $\theta$, $\tau$, and $\phi$.
Here, we discuss why we chose this approach.
To be concrete, let us consider the derivative with respect to $\tau$.
In section \ref{Sec:Adjoint:Adjoint} (specifically, equation \eqref{eq:Loss:Grad:Tau:Naive}), we motivated the adjoint as a way to avoid computing quantities like $\partial_{\tau} x(t)$.
Computing this derivative is possible by solving another IVP forward in time.
We will refer to this approach as the \emph{direct approach}.
Thomas Gronwall originally proposed the direct approach for ODEs \cite{gronwall1919note}.
More recently, Giles and Pierce proposed the direct approach for algebraic equations and PDEs \cite{giles2000introduction}.
Adapting the direct approach to DDEs is possible but computationally infeasible.
Specifically, to compute $\partial_{\theta} x$, $\partial_{\tau}$, and $\partial_{\phi} x$ on $[0, T]$, we need to solve DDEs in $\mathbb{R}^{d \times p}$, $\mathbb{R}^d$, and $\mathbb{R}^{d \times q}$, respectively.
Thus, the number of equations we need to solve (and, therefore, the complexity of obtaining the solutions) to evaluate the gradients grows linearly with the number of parameters.
We initially developed a DDE version of the direct approach but abandoned it in favor of the adjoint approach because the latter is computationally more efficient.
With the adjoint approach, we can use the same adjoint in all gradient computations. 
Thus, we only need to solve one DDE in $\mathbb{R}^d$, which makes the adjoint approach significantly cheaper.
With that said, both approaches do work, though we believe the adjoint approach is superior for practical applications.

\subsection{\texorpdfstring{Why $N_{\tau}$ exists}{}}
\label{Sec:Discussion:Ntau}

\texttt{DDE-Find} uses a second-order Runge-Kutta method to solve the initial value problem, equation \eqref{eq:IVP}.
Our experiments use a time step size $\Delta t = \tau / N_{\tau}$, often with $N_{\tau} = 10$. 
Here, $\tau$ is our model's current guess for the delay (which is not necessarily the true delay).
We do not claim that this choice of $\Delta t$ is optimal.
However, using a $\Delta t$ for which $\tau$ is an integer multiple of $\Delta t$ does have some advantages.
Specifically, to evaluate $x(t - \tau)$, we use the numerical solution from $N_{\tau}$ time steps ago.
If $\tau$ is not a multiple of $\Delta t$, then we would need to use an interpolation of the predicted solution to compute $x(t - \tau)$.
We would also have to update this interpolation after each time step (as we incrementally solve for $x$).
Such a process could become quite expensive and would introduce an extra source of error (as the interpolated solution would not be exact).
Thus, using a $\Delta t$ for which $\tau$ is an integer multiple of $\Delta t$ simplifies the forward pass.
The main downside is that $T$ is often not a multiple of $\Delta t$.
This complication slightly complicates evaluating $\mathcal{L}$.
We discuss how \texttt{DDE-Find} overcomes this challenge in section \ref{Sec:Discussion:Loss}.

\bigskip 

$N_{\tau}$ is a hyperparameter that the user must select.
The gradient for the loss with respect to $\phi$ in equation \eqref{eq:loss} includes an integral on $[0, \tau]$.
We use the trapezoidal rule to approximate this integral (see section \ref{Sec:Discussion:Loss}).
Since we use a step size of $\tau/N_{\tau}$ in the backward pass, we evaluate this integral using just $N_{\tau}$ quadrature points.
If $N_{\tau}$ is small, our quadrature approximation of this integral will have a large error.
However, if $N_{\tau}$ is too large, the forward and backward passes become expensive.
In our experiments, using $N_{\tau}$ between $10$ and $100$ seems to work well, though we do not claim this choice is optimal.
In general, we believe the user should set $N_{\tau}$ as large as is feasible on their machine.

\subsection{\texorpdfstring{Evaluating $\mathcal{L}$}{}}
\label{Sec:Discussion:Loss}

\texttt{DDE-Find} approximates the integral loss, $\int_{0}^{T} \ell(x(t))\ dt$, using a quadrature rule (specifically, the trapezoidal rule).
To evaluate this quadrature rule, we split the interval $[0, T]$ into evenly sized sub-intervals and then evaluate the integrand at the boundaries of those sub-intervals.
Thus, to evaluate the approximation, \texttt{DDE-Find} needs to know the predicted and target trajectory at a finite set of points in $[0, T]$.
Since the target trajectory interpolates the data, we can evaluate it anywhere.
However, because we use a numerical DDE solve to find the predicted solution, we only know the predicted trajectory at $t = 0, \Delta t, 2\Delta t, \ldots, N_{step} \Delta t$.
In general, these times are not the quadrature points at which we need to evaluate the integrand.

\bigskip

We could approximate the predicted trajectory at the quadrature points by interpolating the numerical solution.
Packages such as \texttt{Scipy} \cite{virtanen2020scipy}, which we use to compute target trajectory, can do this. 
However, our implementation of \texttt{DDE-Find} uses \texttt{Pytorch} \cite{paszke2019pytorch} and its automatic differentiation capabilities, highly optimized operations, and its array of optimizers.
We implemented our adjoint method in a custom \texttt{torch.autograd.Function} class that implements the adjoint approach in section \ref{Sec:Method} to tell  \texttt{Pytorch} how to differentiate the loss with respect to the parameters.
To make \texttt{Pytorch} use our adjoint approach, however, \texttt{Pytorch} needs to construct a computational graph from the DDE solve in the forward pass to $\mathcal{L}$.
In general, to use a third-party interpolation tool, like \texttt{Scipy}, we would need to convert the predicted trajectory to a \texttt{Numpy} array and then pass the result to the third-party tool to obtain the interpolation.
By doing so, we lose the computational graph connection between the predicted trajectory and loss, meaning that \texttt{Pytorch} will not use our adjoint approach.
This deficiency rules out using third-party interpolation tools on the target trajectory.

\bigskip 

Notably, \texttt{DDE-Find} does not need to know the precise value of the loss to compute its gradients.
Thus, while we could implement interpolation strategies using torch operations, we chose not to.
During the backward pass, we use an interpolation of the predicted trajectory to evaluate the predicted trajectory at arbitrary points in $[0, T]$, solve the adjoint equation, and compute the gradients of the loss.
This approach only relies on being able to evaluate the predicted trajectory, not on being able to report the exact value of the loss.

\bigskip 

Given this insight, we decided to abandon evaluating the loss exactly.
Specifically, \texttt{DDE-Find} uses a quadrature approximation of $\int_{0}^{N_{step} \Delta t} \ell(x(t))\ dt$ to approximate the running loss.
We use sub-intervals of width $\Delta t = \tau / N_{\tau}$. 
This choice means that we only need to know the predicted trajectory at $t = 0, \Delta t, 2\Delta t, \ldots$, which is precisely what our numerical solution gives us.
Thus, we can evaluate the integral loss directly using the numerical approximation to the predicted trajectory.
Likewise, for the terminal loss, \texttt{DDE-Find} evaluates $G(x(N_{step} \Delta t))$, not $G(x(T))$.
In general, since both $G$ and $x$ are continuous, and $T \approx N_{step} \Delta t$, we expect that $G(x(N_{step} \Delta t)) \approx G(x(T))$.
This approach means we can evaluate the loss without interpolating the predicted trajectory, but \texttt{DDE-Find} reports an erroneous loss.
Further, with this approach, \texttt{Pytorch} constructs a computational graph from the forward pass to the loss, meaning it triggers our adjoint code during the backward pass.
Finally, because we only need to know the predicted trajectory at $0, \Delta t, \ldots, N_{\tau} \Delta t$ to compute the gradients of the loss, \texttt{DDE-Find} still computes an accurate adjoint and accurate gradients of $\mathcal{L}$.

\subsection{\texorpdfstring{The Challenge of Learning $\phi$ }{}}
\label{Sec:Discussion:phi}

Throughout our experiments in section \ref{Sec:Experiments}, \texttt{DDE-Find} struggles to learn some components of $\phi$.
This limitation is especially apparent when learning the $A$ and $\omega$ components of periodic boundary conditions, equation \eqref{Eq:X0:Periodic}. 
Here, we take a closer look at this phenomenon.
Specifically, we argue that our results are not due to a problem with \texttt{DDE-Find}; instead, they are a fundamental limitation of any algorithm that attempts to learn $\theta, \tau$, and $\phi$ by minimizing the loss function, equation \eqref{eq:loss}.
To summarize what follows, changes in some of the components of $\phi$ have little to no impact on the predicted trajectory, meaning the gradient of $\mathcal{L}$ with respect to these coefficients is small or, in the case of the HIV model in section \ref{Sec:Experiments:HIV}, identically zero.

\bigskip 

To aide in this discussion, we re-state the expression for $\nabla_{\phi} \mathcal{L}$ in equation \eqref{eq:gradients}:
\begin{equation}
\label{eq:loss:phi}
\begin{aligned}
    \nabla_{\phi} \mathcal{L}\left( x \right) =& - \left[ \partial_{\phi} X_0(0, \phi) \right]^T \lambda(0) \\
    &- \int_{0}^{\tau} \left[ \partial_\phi X_0(t - \tau, \phi) \right]^T \left[\partial_{y} F\left( x(t), X_0(t - \tau, \phi), \tau, t, \theta \right)\right]^T \lambda(t) \ dt
\end{aligned}
\end{equation}
Let's focus on the first portion of this expression: $\left[ \partial_{\phi} X_0(0, \phi) \right]^T \lambda(0)$.
Notice that for $X_0(t, \phi) = A \sin(\omega t) + b$, we have
\begin{equation*}
\begin{aligned}
    \frac{\partial}{\partial A} X_0(t, \phi) &= \sin(\omega t) \\
    \frac{\partial}{\partial \omega} X_0(t, \phi) &= A t cos(\omega t),
\end{aligned}
\end{equation*}
both of which are zero at $t = 0$. 
Further, by continuity, we expect both derivatives will be small (but non-zero) for $t$ close to zero. 
If $\omega \tau < \pi/2$, then we expect $\partial X_0 / \partial A$ and $\partial X_0 / \partial \omega$ will be small for most of the interval $[0, \tau]$, meaning we expect $\mathcal{L}$ to be insensitive to changes in $A$ or $\omega$.
In this case, we expect \texttt{DDE-Find} to have trouble recovering these parameters from noisy measurements; the noise may shift the shallow local minimum of $\mathcal{L}$ with respect to these parameters.
This observation matches our results.
Specifically, \texttt{DDE-Find} struggles to identify $A$ and $\omega$ in the Logistic decay experiments (where $\tau \omega$ is small), but can reliably identify $\omega$ in the ENSO experiments (where $\tau \omega$ is larger). 

\bigskip 

An extreme version of this phenomenon happens in the HIV experiments.
Specifically, the $T^*$ and $V_{NI}$ components of $A$ and $\omega$ do not change from their initial value.
There is a good reason for this: the loss doesn't depend on these components of the initial condition.
To see this, let us first consider $\nabla_{\phi} \mathcal{L}$. 
In this case, the only component of the DDE that depends on the delayed state is the $T^*$ equation, which depends on $V_{I}(t - \tau)$. 
Thus, the $V_{I}, V_{NI}$ component is the only non-zero cell of $\partial_y F$.
Thus, $\partial_y F( x(t), X_0(t - \tau, \phi), \tau, t, \theta )^T \lambda(t)$ has zeros in its $T^*$ and $V_{NI}$ entries.
By inspection, the rows of $\left[ \partial_\phi X_0(t - \tau, \phi) \right]^T$ corresponding to $\partial_{A[T^*]} X_0$,  $\partial_{A[V_{NI}]} X_0$, $\partial_{\omega[T^*]} X_0$, and $\partial_{\omega[V_{NI}]} X_0$ have a zero in their $V_{I}$ component.
Therefore, the product 
$$\left[ \partial_\phi X_0(t - \tau, \phi) \right]^T \left[\partial_{y} F\left( x(t), X_0(t - \tau, \phi), \tau, t, \theta \right)\right]^T \lambda(t)$$
has zeros in the components corresponding to $A[T^*]$, $A[V_{VI}]$, $\omega[T^*]$, and $\omega[V_{NI}]$. 
However, since $- \left[ \partial_{\phi} X_0(0, \phi) \right]^T \lambda(0)$
is also zero for these components (as discussed above), the components of $\nabla_{\phi} \mathcal{L}$ corresponding 
to $A[T^*]$, $A[V_{NI}]$, $\omega[T^*]$ and $\omega[V_{NI}]$ are zero.
In other words, \texttt{DDE-Find} never updates these parameters from their initial values because the derivative of the loss with respect to these parameters is identically zero.

\bigskip 

We can take this analysis further, however. 
In a DDE, the true trajectory depends on the initial condition function through the delay term.
To see this, consider the general DDE IVP we consider in this paper:
\begin{equation*}
\begin{aligned}
    \dot{x}(t) &= F\left(x(t), x(t - \tau), t, \tau, \theta \right) \qquad && t \in [0, T] \\
    x(t) &= X_0(t, \phi) && t \in [-\tau, 0]
\end{aligned}
\end{equation*}
For $t \in [0, \tau]$, we have $\dot{x}(t) = F(x(t), X_0(t - \tau, \phi), t, \tau, \theta)$.
Thus, in general, the rate of change of $x$ depends on the initial condition when $t \in [0, \tau]$.
However, in the HIV model, $V_{I}(t - \tau)$ is the only delay term that appears in the IVP, equation \eqref{eq:IVP:HIV}.
Further, we know that $X_0(0, \phi)$ does not depend on $A$ or $\omega$.
These results the predicted trajectory does not depend on the $T^*$ and $V_{NI}$ components of $A$ and $\omega$; changing these four coefficients will not change the predicted trajectory.

\bigskip 

Given this insight, it is no surprise that \texttt{DDE-Find} can not recover these components of $\phi$.
We learn $\theta$, $\tau$, and $\phi$ by searching for a combination of parameters whose predicted trajectory matches the true one.
Throughout all of our experiments, \texttt{DDE-Find} accomplishes this task admirably. 
However, we don't really care about matching the predicted and true trajectories.
Instead, we care about learning $\theta$, $\tau$, $\phi$; we hope that minimizing the loss, equation \eqref{eq:loss}, will yield these coefficients as a by-product, though the analysis above shows that this is not always the case.
\texttt{DDE-Find} does what we ask, but that doesn't necessarily mean it's doing what we want.

\bigskip

Critically, this deficiency is not specific to \texttt{DDE-Find}.
Any algorithm that attempts to learn $\theta$, $\tau$, and $\phi$ by minimizing a loss function similar to equation \eqref{eq:loss} would suffer from this issue.
As far as the author is aware, every existing algorithm for learning DDEs from data (including \cite{sandoz2023sindy}, \cite{stephany2024learning}, \cite{anumasa2021delay}, and \cite{zhu2021neural}) uses an analogous approach and would suffer from the same deficiency.
The universality of this limitation in existing approaches suggests a potential future research direction: Designing a new loss function.
Our loss function (and those in previous work) uses the $L^1$, $L^2$, or $L^{\infty}$ norms of the difference between the predicted and target trajectories.
We believe that designing a loss function with a unique (or, at the very least, steep) minimizer at the true parameter values (without imposing unrealistic assumptions on the data) represents an important future research direction.

\subsection{\texorpdfstring{Using Neural Networks for $F$ and $X_0$}{}}
\label{Sec:Discussion:Neural}

Our implementation of \texttt{DDE-Find} allows $F$ and $X_0$ to be any \texttt{torch.nn.Module} object, including deep neural networks.
Using deep neural networks, however, introduces some additional problems.
One issue is initialization.
Suppose we use a neural network for $F$ and initialize its weight matrices using \emph{Glorot initialization} \cite{glorot2010understanding} (one of the standard initialization schemes for deep neural networks).
In many cases, the map $t \to F(x(t), x(t - \tau), t, \tau, \theta)$ has the same sign on all or most of $[0, T]$, which causes the initial predicted solution to grow exponentially.
In some instances, the predicted solution exceeds the 32-bit floating point limit before reaching $t = T$.
In such cases, \texttt{DDE-Find} fails. 
This result isn't surprising; The authors of Glorot initialization designed their approach to address the vanishing gradient problem, not to make a network that can act as the right-hand side of a DDE.
We believe that better, application-specific initialization schemes are necessary to make using deep neural networks for $F$ or $X_0$ practical.

\bigskip 

\begin{figure}[H]
    \centering
    \includegraphics[width=0.75\textwidth]{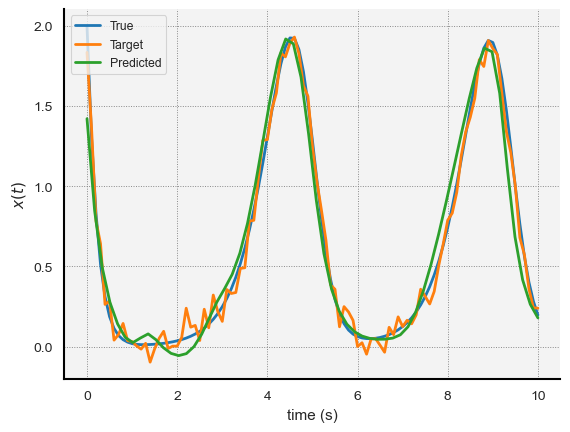}
    \caption{The true, target, and predicted trajectory for the Logistic model when we use a neural network for $F$.}
    \label{fig:Logistic:Neural}
\end{figure}

Further, the large number of parameters in a neural network can over-parameterize $F$, leading to unexpected results.
To make this concrete, we return to the Logistic Delay equation and use a neural network for $F$.
The network for $F$ has two hidden layers with ten neurons each.
We train the network for $1000$ epochs with a learning rate of $0.03$.
Figure \ref{fig:Logistic:Neural} shows the true, target, and predicted trajectories after training.
Notice that the predicted trajectory closely matches the true one.
However, \texttt{DDE-Find} learns $\tau = 1.71964$, while the true $\tau$ value is $1.0$.
Thus, \texttt{DDE-Find} learns a $(\theta, \tau, \phi)$ combination that produces a trajectory that matches the true trajectory but identifies the wrong delay value in the process.
\texttt{DDE-Find} does what we ask it to (minimize the loss) but does not give us the answer we want ($\tau = 1.0$).
Thus, there is a DDE with a delay of $\approx 1.72$ whose solution is nearly identical to the true trajectory.
In other words, the true solution satisfies multiple DDEs with multiple delays.
This result somewhat complicates identifying \emph{the} parameters, delay, and initial conditions that engender the true solution.
An alternative loss function, as discussed in section \ref{Sec:Discussion:Loss}, might mitigate this non-uniqueness problem.

\bigskip 

Nonetheless, if $F$ or $X_0$ are neural networks, \texttt{DDE-Find} can learn $\theta, \tau$, and $\phi$ that engender a predicted trajectory that matches the true one.
We believe using a neural network for $F$ or $X_0$ is most appropriate when the user is more interested in making predictions using the underlying (hidden) dynamics rather than learning a specific set of parameters (system identification).
In the latter case, structured models, such as the ones we consider in our experiments, are more appropriate.

\section{Conclusion}
\label{Sec:Conclusion}

In this paper, we introduced \texttt{DDE-Find}, an algorithm that learns a DDE from a noisy measurement of its solution.
To do this, \texttt{DDE-Find} a set of parameters, $(\theta, \tau, \phi)$, such that if we solve equation \eqref{eq:IVP} using those parameters, the corresponding predicted trajectory, $x$, closely matches the data.
To learn these parameters, \texttt{DDE-Find} minimizes a loss function that depends on the predicted trajectory and the data.
\texttt{DDE-Find} uses an adjoint approach to compute the gradients of the loss with respect to $\theta$, $\tau$, and $\phi$.
It then uses a gradient descent approach to optimize the parameter values.
\texttt{DDE-Find} is efficient, with the gradient computations requiring the same order of work as the forward pass (solving the DDE).
Further, through a series of experiments, we demonstrate that \texttt{DDE-Find} is effective at learning $\theta$, $\tau$, and $\phi$ from noisy measurements of several DDE models. 
\texttt{DDE-Find} is a general-purpose algorithm for learning DDEs from data.
We believe it could be an important tool in discovering new DDE models from scientific data.

\printbibliography
\appendix
\section{Proof of Theorem \ref{Theorem:1}}
\label{Sec:Proof}

In this section, we provide a proof of Theorem \ref{Theorem:1}.
Our proof is based on the adjoint state method \cite{pontryagin2018mathematical} and is loosely inspired by the proof in \cite{ayed2019learning}. 
Our approach mirrors the one in \cite{stephany2024learning}, which proved a special case of theorem \ref{Theorem:1}. 
Our proof uses myriad results on Frechet differentiation.
However, proofs of all the results we use can be found in \cite{munkres2018analysis}.
Before we formally prove Theorem \ref{Theorem:1}, we provide some notation, motivation, and key insights.

\subsection{Derivative Notation}
\label{Sec:Proof:Notation}

Let us begin by clarifying our derivative notation.
This proof involves derivatives of functions of the form $a : \mathbb{R}^n \to \mathbb{R}^m$, $b : \mathbb{R}^n \to \mathbb{R}$, and $c : \mathbb{R} \to \mathbb{R}$.
Different authors use various notations for these derivatives.
Here, we adopt the notation used in \cite{stephany2024learning}.
For completeness, we restate the notation below.

\bigskip

We write $\dot{h}(t)$ for the time derivative $h$, regardless of whether $h$ is real, vector, or matrix-valued.
The notation below applies to derivatives with respect to variables other than time.
For functions of the form $a : \mathbb{R}^n \to \mathbb{R}^m$, we let $\partial_x a(x) \in \mathbb{R}^{m \times n}$ denote the matrix of the (Frechet) derivative of $a$ at $x$ \cite{munkres2018analysis}.
If $n = 1$, we usually interpret the $m \times 1$ matrix $\partial_x a(x)$ as a vector in $\mathbb{R}^m$. 
Next, for functions of the form $b : \mathbb{R}^n \to \mathbb{R}$, we let $\nabla b(y) \in \mathbb{R}^n$ denote the gradient of $b$ at $y \in \mathbb{R}^n$.
Finally, for functions of the form $c : \mathbb{R} \to \mathbb{R}$, we let $\partial c(z) / \partial z \in \mathbb{R}$ denote the (partial) derivative of $c$ with respect to $z$.

\bigskip 

In some cases, the above notation can be ambiguous.
Specifically, consider the case of a vector-valued function, $a$, of two variables, $x$ and $y$.
Assume, however, that $y$ is a function of $x$ and consider the expression $\partial_{x} a\left(x, y(x) \right)$. 
It's unclear if this refers to the Frechet derivative of the map $(x, y) \to a(x, y)$ $a$ with respect to its first argument, or of the map $x \to a(x, y(x))$ (which itself is a composition of the map $x \to (x, y(x))$ with the map $(x, y) \to a(x, y)$).
Thus, whenever it is ambiguous, we use the notation $\partial_{x} a\left(x, y(x)\right)$ to denote the derivative of the map $(x, y) \to a(x, y)$ with respect to its first argument at $(x, y(x))$ and introduce the notation $D_{x} H\left(x , y(x) \right)$ to denote the (total) Frechet derivative of the map $x \to a(x, y(x))$ at $(x, y(x))$. 
We adopt the same $D$ notation for functions real-valued functions, $b$, with multiple arguments.
That is, $D_{x} b(x, y(x))$ denotes the (total) Frechet derivative of the map $x \to b(x, y(x))$, while $\nabla_{x} b(x, y(x))$ denotes the Frechet derivative of the map $(x, y) \to b(x, y)$ with respect to its first argument at $(x, y(x))$.

\bigskip 

Finally, when writing derivatives of $F$ (in equation \eqref{eq:IVP}) we use the notation $\partial_{x} F$ and $\partial_{y} F$ to denote the Frechet derivatives of $F$ with respect to its first and second arguments, respectively.

\subsection{The Lagrangian}
\label{Sec:Proof:Lagrangian}

To motivate our approach, we begin with a critical observation.
Suppose that $x : [-\tau, T] \to \mathbb{R}^d$ is a solution to the IVP, equation \eqref{eq:IVP}, for a specific $\tau > 0$, $\theta \in \mathbb{R}^p$, and $\phi \in \mathbb{R}^q$. 
Then, we must have
\begin{equation}
\label{eq:IVP:dot_x}
\begin{aligned}
    \dot{x}(t) &= F\left( x(t), x(t - \tau), \tau, t, \theta\right) &&  t \in [0, T] \\
    x(t) &= X_0\left(t, \phi\right) &&  t \in [-\tau, 0).
\end{aligned}
\end{equation}
Specifically,
$$\dot{x}(t) - F\big( x(t), x(t - \tau), \tau, t, \theta\big) = 0 \qquad t \in [0, T].$$
With this in mind, let $\lambda : [0, T] \to \mathbb{R}^d$ be an arbitrary function of class $C^1$.
Then, we must have
$$\left\langle \lambda(t),\ \dot{x}(t) - F\big( x(t), x(t - \tau), \tau, t, \theta\big) \right\rangle = 0 \qquad t \in [0, T],$$
which means that
\begin{equation}
\label{eq:Lagrangian:Integral}
    \int_{0}^{T} \left\langle \lambda(t),\ \dot{x}(t) - F\big( x(t), x(t - \tau), \tau, t,\theta\big) \right\rangle\ dt = 0.
\end{equation}
Critically, equation \eqref{eq:Lagrangian:Integral} must hold irrespective of the value of $\lambda$, $\tau$, $\theta$, or $\phi$. 
Because $x$ is defined as the solution to equation \eqref{eq:IVP}, changing $\tau$, $\theta$, or $\phi$ causes $x$ to correspondingly changes such that equation \eqref{eq:Lagrangian:Integral} still holds.
Notably, this means that $x$ is implicitly a function of $\tau$, $\theta$, and $\phi$. 
At its core, the proof we present later in this section holds because of equation \eqref{eq:Lagrangian:Integral}.

\bigskip 

Based on these observations, we introduce the Lagrangian, $L : C^1\left[ [0, T], \mathbb{R}^d \right] \times \mathbb{R}^p \times (0, \infty) \times \mathbb{R}^q \to \mathbb{R}$, as follows:
\begin{equation}
\label{eq:Lagrangian}
    L\left(\lambda, \theta, \tau, \phi\right) = \mathcal{L}\left(x\right) + \int_{0}^{T} \left\langle\lambda(t),\ \dot{x}(t) - F\left(x(t), x(t - \tau), \tau, t, \theta\right) \right\rangle\ dt.
\end{equation}
Here, $x : [-\tau, T] \to \mathbb{R}^d$ is the solution to equation \eqref{eq:IVP}.
Notice that this expression does not explicitly depend on $\theta$ or $\phi$. 
Rather, $L$ depends on $\theta$ and $\phi$ implicitly through $x$ (and, as we will see, $\lambda$). 
Finally, because of equation \eqref{eq:Lagrangian:Integral}, we have the following identity 
\begin{equation}
\label{eq:Lagrangian:Identity}
    L\left(\lambda, \theta, \tau, \phi\right) = \mathcal{L}\left(x\right) \qquad \forall\ \lambda, \theta, \tau, \phi.
\end{equation}
In particular, this means that if $L$ and $\mathcal{L}$ are Frechet differentiable with respect to $\tau$, $\theta$, and $\phi$ (a property we will justify in the forthcoming proof) then 
\begin{equation}
\label{eq:Lagrangian:Identity:Gradient}
\begin{aligned}
    \nabla_{\theta} L\left(\lambda, \theta, \tau, \phi\right) &= \nabla_{\theta} \mathcal{L}(x) \\
    \frac{\partial}{\partial \tau} L \left(\lambda, \theta, \tau, \phi\right) &= \frac{\partial}{\partial \tau} \mathcal{L}(x) \\
    \nabla_{\phi} L\left(\lambda, \theta, \tau, \phi\right) &= \nabla_{\phi} \mathcal{L}(x).
\end{aligned}
\end{equation}
In particular, this means that if we can find a $\lambda$ which simplifies evaluating the gradients of $L$, we can use it to compute the gradients of the loss, $\mathcal{L}$.
This observation forms the basis of our approach.

\subsection{Proof of Theorem \ref{Theorem:1}}
\label{Sec:Proof:Thorem1}

For ease of reference, we begin by restating Theorem \ref{Theorem:1}:

\addtocounter{theorem}{-1}
\begin{theorem}
    Let $T, \tau > 0$, $\theta \in \mathbb{R}^{p}$, and $\phi \in \mathbb{R}^{q}$.
    Let $x : [-\tau, T] \to \mathbb{R}^d$ solve the initial value problem in equation \eqref{eq:IVP}.
    Further, suppose that assumptions \ref{assumption:FGell}, \ref{assumption:X0}, and \ref{assumption:x} hold. 
    Let $\lambda: [0, T] \rightarrow \mathbb{R}^n$ satisfy the following:
    \begin{equation*}
    \begin{aligned}
        \dot{\lambda}(t) & = \nabla_x \ell \left( x(t) \right) - \left[ \partial_x F\left(x(t), x(t - \tau), \tau, t, \theta\right) \right]^T \lambda(t) \\
        &\quad - \mathbb{1}_{t < T - \tau}(t) \left[ \partial_y F \left( x(t + \tau), x(t), \tau, t + \tau, \theta\right) \right]^T \lambda\left(t + \tau \right) \qquad \ t \in [0, T]  \\
        \lambda(T) &= -\nabla_x G \left(x(T)\right)
    \end{aligned}
    \end{equation*}
    Then, the derivatives of the loss function, $\mathcal{L}$, with respect to $\theta, \tau$ and $\phi$ are:
    \begin{equation*}
    \begin{aligned}
        \nabla_{\theta} \mathcal{L}\left( x \right) =& - \int_0^T  \left[ \partial_{\theta} F (x(t), x(t - \tau), \tau, t, \theta) \right]^T \lambda(t) \ dt \\
        \frac{\partial}{\partial\tau} \mathcal{L}\left( x \right) =& \int_0^{T - \tau} \left\langle \left[ \partial_y F (x(t + \tau), x(t), \tau, t + \tau, \theta) \right]^T \lambda(t + \tau),\ \dot{x}(t) \right\rangle\ dt \\
        -& \int_{0}^{T} \left\langle \partial_{\tau} F(x(t), x(t - \tau), \tau, t, \theta),\ \lambda(t) \right\rangle \ dt \\
        \nabla_{\phi} \mathcal{L}\left( x \right) =& - \left[ \partial_{\phi} X_0(0, \phi) \right]^T \lambda(0) \\
        - &\int_{0}^{\tau} \left[ \partial_\phi X_0(t - \tau, \phi) \right]^T \left[\partial_{y} F\left( x(t), X_0(t - \tau, \phi), \tau, t, \theta \right)\right]^T \lambda(t) \ dt
    \end{aligned}
    \end{equation*}
\end{theorem}
For a proof of the $\nabla_{\theta} \mathcal{L}$ equation, we refer the reader to \cite{stephany2024learning}.
That paper solves a special case of Theorem \ref{Theorem:1}, though their proof of the $\theta$ gradient does not rely on their stronger assumptions.
Thus, we can use the exact same proof to arrive at the $\theta$ gradient.
In the next two subsections, we prove the expressions for $\partial \mathcal{L} / \partial \tau$ and $\nabla_{\phi} \mathcal{L}$ (which, as far as the authors of this paper are aware, has not appeared in the literature).

\subsubsection{\texorpdfstring{$\frac{\partial}{\partial \tau} \mathcal{L}$}{}}
\label{Sec:Proof:dLdtau}

To begin, fix $(\theta, \tau, \phi) \in \mathbb{R}^p \times (0, \infty) \times \mathbb{R}^q$.
In this section, we will denote the solution to equation \eqref{eq:IVP}, $x$, by $x_{\tau}$ to make its dependence on $\tau$ explicit.
Further, in general, we will assume that $\lambda$ depends on $\tau$.
To make this dependence clear, we will denote $\lambda$ by $\lambda_{\tau}$.
We restrict our attention to $\lambda_{\tau}$ for which the map $(\tau, t) \to \lambda_{\tau}(t)$ is of class $C^1$.

\bigskip 

We can now begin the proof.
First, we will argue that $L$ is differentiable with respect to $\tau$. 
We will then compute the derivative of $L$ with respect to $\tau$, introduce the adjoint, and then derive our final result.
To begin, we will first show that $\mathcal{L}$ is differentiable with respect to $\tau$.
By assumption, the map $\tau \to x_{\tau}(T)$ is of class $C^2$. 
Further, the map $y \to G(y)$ is of class $C^1$.
Thus, both maps must be (Frechet) differentiable.
By assumption, the map $\tau \to x_{\tau}(t)$ is differentiable for each $t \in [0, T]$.
Therefore, by the chain rule, the map $\tau \to G\left( x_{\tau}(T) \right)$ is differentiable with
$$\frac{\partial}{\partial \tau}  G(x_{\tau}(t)) = \left\langle \nabla_{x} G\left(x_{\tau}(T)\right),\  \partial_{\tau} x_{\tau}(T) \right\rangle.$$
Similarly, for each $t \in [0, T]$, the map $\tau \to \ell(x_{\tau}(t))$ is differentiable with 
$$\frac{\partial}{\partial \tau}  \ell(x_{\tau}(t)) = \left\langle \nabla_{x} \ell(x_{\tau}(t)),\ \partial_{\tau} x_{\tau}(t) \right\rangle.$$
Critically, the components of the latter derivative are continuous on $[0, T]$.
Therefore, we can swap the order of differentiation and integration.
In particular, the map $\tau \to \int_{0}^{T} \ell(x_{\tau}(t))\ dt$ must be differentiable with derivative
$$\int_{0}^{T} \left\langle \nabla_{x} \ell(x_{\tau}(t)),\ \partial_{\tau} x_{\tau}(t) \right\rangle\ dt.$$
Combining these results, we can conclude that the map $\tau \to \mathcal{L}(x_{\tau})$ is differentiable with
\begin{equation}
\label{eq:Loss:Grad:Tau}
    \frac{\partial}{\partial \tau} \mathcal{L}\left( x_{\tau} \right) = \left\langle \nabla_{x} G\left( x_{\tau}(T) \right),\ \partial_{\tau} x_{\tau}(T) \right\rangle + \int_{0}^{T} \left\langle \nabla_{x} \ell\left( x_{\tau} (t) \right),\ \partial_{\tau} x_{\tau}(t) \right\rangle\ dt.
\end{equation}
With this established, we can move onto the full Lagrangian.
To begin, fix $t \in [0, T]$ and consider the map $\tau \to \left\langle \lambda_{\tau}(t),\  \dot{x}_{\tau}(t) - F\left(x_{\tau}(t), x_{\tau}(t - \tau), \tau, t, \theta\right) \right\rangle$.
Let's focus on the map 
$$\tau \to F\left(x_{\tau}(t), x_{\tau}(t - \tau), \tau, t, \theta\right).$$
We use the chain rule to show that this map is differentiable.
To begin, the maps $\tau \to x_{\tau}(t)$, $\tau \to \tau$, $\tau \to t$, and $\tau \to \theta$ are differentiable (the first by assumption, the second because it is the identity, and the third because it is a constant).
What remains is the map $\tau \to x_{\tau}(t - \tau)$. 
This map depends on $\tau$ explicitly (through the argument $t - \tau$) and implicitly (since changing $\tau$ changes $x_{\tau}$. 
However, notice that the map $\tau \to (\tau, t - \tau)$ is of class $C^2$.
Further, by assumption, the map $(t, \tau) \to x_{\tau}(t)$ is of class $C^2$.
Therefore, by the chain rule, the map $\tau \to x_{\tau}(t - \tau)$ (which is the composition of the two maps we just considered) is of class $C^2$ with derivative 
$$\partial_{\tau} x_{\tau}(t - \tau) - \dot{x}_{\tau}(t - \tau).$$
Thus, the component functions of the maps $\tau \to x_{\tau}(t)$, $\tau \to x_{\tau}(t - \tau)$, $\tau \to \tau$, $\tau \to t$, and $\tau \to \theta$ have continuous partial derivatives.
Therefore, we can conclude that the component functions of the map $\tau \to \left( x_{\tau}(t), x_{\tau}(t - \tau), \tau, t, \theta \right)$ have continuous partial derivatives.
Hence, this map must be of class $C^1$.
Since $F$ is also of class $C^1$, the chain rule tells us that the map $\tau \to F\left( x_{\tau}(t), x_{\tau}(t - \tau), \tau, t, \theta \right)$ is of class $C^1$.
Therefore, this map is differentiable with 
\begin{equation}
\label{eq:F:Grad:Tau}
    D_{\tau} F(t) = \partial_{x} F(t) \partial_{\tau} x_{\tau}(t) + \partial_{y} F(t) \left( \partial_{\tau} x_{\tau}(t - \tau) - \dot{x}_{\tau}(t - \tau) \right) + \partial_{\tau} F(t).
\end{equation}
Here, for brevity, we let $F(t)$ be an abbreviation of $F\left( x_{\tau}(t), x_{\tau}(t - \tau), \tau, t, \theta \right)$.
Since the map $(t, \tau) \to x_{\tau}(t)$ is of class $C^2$, the map $\tau \to \dot{x}_{\tau}(t)$ must be differentiable as well.
Further, equality of mixed partials tells us that 
$$\partial_{\tau} \dot{x}_{\tau}(t) = \dot{\partial_{\tau} x_{\tau}}(t).$$
Finally, since the map $\tau \to \lambda_{\tau}(t)$ is of class $C^1$, the component functions of both arguments of the inner product $\left\langle \lambda_{\tau}(t),\ \dot{x}_{\tau}(t) - F\left(x_{\tau}(t), x_{\tau}(t - \tau), \tau, t, \theta\right) \right\rangle$ have continuous partial derivatives.
Since this inner product is a linear combination of the products of the components of these arguments, we can conclude that the map $\tau \to \left\langle \lambda_{\tau}(t),\ \dot{x}_{\tau}(t) - F\left(x_{\tau}(t), x_{\tau}(t - \tau), \tau, t, \theta\right) \right\rangle$ is differentiable with 
\begin{equation*}
\begin{aligned}
    \frac{\partial}{\partial \tau} \left\langle x_{\tau}(t),\ \dot{x}_{\tau}(t) - F(t) \right\rangle &= \\
    \big\langle \partial_{\tau} \lambda_{\tau}(t),\ \dot{x}_{\tau}(t) &- F\left( x_{\tau}(t), x_{\tau}(t - \tau), \tau, t, \theta \right) \big\rangle + \\
    \Big\langle \lambda_{\tau}(t),\ \dot{\partial_{\tau}x}(t) &- \left( \partial_{x} F(t)\right) \partial_{\tau}x_{\tau}(t) - \left( \partial_{y} F(t) \right) \left( \partial_{\tau} x_{\tau}(t - \tau) - \dot{x}_{\tau}(t - \tau) \right) - \partial_{\tau} F(t) \Big\rangle. 
\end{aligned}
\end{equation*}
Notably, since $\dot{x}_{\tau}(t) = F\left( x_{\tau}(t), x_{\tau}(t - \tau), \tau, t, \theta \right)$, we must have 
$$\big\langle \partial_{\tau} \lambda_{\tau}(t),\ \dot{x}_{\tau}(t) - F\left( x_{\tau}(t), x_{\tau}(t - \tau), \tau, t, \theta \right) \big\rangle = 0,$$
which means that 
\begin{equation}
\label{eq:Integrand:Grad:Tau}
\begin{aligned}
    \frac{\partial}{\partial \tau} \left\langle x_{\tau}(t),\ \dot{x}_{\tau}(t) - F(t) \right\rangle &= \left\langle \lambda_{\tau}(t),\ \dot{\partial_{\tau}x}(t) \right\rangle - \Big\langle \lambda_{\tau}(t),\ \left[ \partial_{x} F(t) \right] \partial_{\tau}x_{\tau}(t) \Big\rangle \\
    &- \Big\langle \lambda_{\tau}(t),\ \left[ \partial_{y} F(t) \right] \big( \partial_{\tau} x_{\tau}(t - \tau) - \dot{x}_{\tau}(t - \tau) \big) \Big\rangle - \Big\langle \lambda_{\tau}(t),\ \partial_{\tau} F(t) \Big\rangle.
\end{aligned}
\end{equation}
Since we chose $t$ arbitrarily, this must hold for each $t \in [0, T]$. 
Further, by inspection, both arguments of the inner product on the right-hand side of equation \eqref{eq:Integrand:Grad:Tau} are continuous functions of $t$.
Therefore, the map $t \to \frac{\partial}{\partial \tau} \left\langle x_{\tau}(t),\ \dot{x}_{\tau}(t) - F(t) \right\rangle$ must be continuous.
Therefore, we conclude that map $\tau \to \int_{0}^{T} \left\langle \lambda_{\tau}(t),\ \dot{x}_{\tau}(t) - F\left( x_{\tau}(t), x_{\tau}(t - \tau), \tau, t, \theta \right) \right\rangle\ dt$ is differentiable.
Further, the derivative of the integral must be equal to the integral of the derivative.
Hence, we must have
\begin{equation}
\label{eq:Integral:Grad:Tau}
\begin{aligned}
    \frac{\partial}{\partial \tau} \int_{0}^{T} \Big\langle \lambda_{\tau}(t),\ &\dot{x}_{\tau}(t) - F(t) \Big\rangle\ dt = \\
    \int_{0}^{T} &\left\langle \lambda_{\tau}(t),\ \dot{\partial_{\tau}x}(t) \right\rangle - \Big\langle \lambda_{\tau}(t),\ \left[ \partial_{x} F(t) \right] \partial_{\tau}x_{\tau}(t) \Big\rangle \\
    - &\Big\langle \lambda_{\tau}(t),\ \left[ \partial_{y} F(t) \right] \big( \partial_{\tau} x_{\tau}(t - \tau) - \dot{x}_{\tau}(t - \tau) \big) \Big\rangle - \Big\langle \lambda_{\tau}(t),\ \partial_{\tau} F(t) \Big\rangle\ dt.
\end{aligned}
\end{equation}
By combining equations \eqref{eq:Loss:Grad:Tau} and \eqref{eq:Integral:Grad:Tau}, we can conclude that the map $\tau \to L\left(\lambda_{\tau}, \theta, \tau, \theta \right)$ is differentiable with
\begin{equation}
\label{eq:L:Grad:Tau:Initial}
\begin{aligned}
    D_{\tau} L\left( \lambda_{\tau}, \theta, \tau, \phi \right) = \int_0^T &\left\langle \nabla_x \ell(x_{\tau}(t)),\ \partial_{\tau} x_{\tau}(t) \right\rangle + \color{teal}\left\langle \lambda_{\tau}(t),\ \dot{\partial_{\tau} x_{\tau}}(t) \right\rangle \color{black} - \left\langle \lambda_{\tau}(t),\ \partial_{\tau} F(t) \right\rangle \\
    - &\left\langle \lambda_{\tau}(t),\ \left[ \partial_{x} F(t) \right] \partial_{\tau} x_{\tau}(t) + \color{violet}\left[ \partial_{y} F(t) \right]\partial_{\tau} x_{\tau}(t - \tau) \color{black} -  \color{orange}\left[ \partial_{y} F(t) \right] \dot{x}_{\tau}(t - \tau)\color{black} \right\rangle\ dt \\
    + &\big\langle \nabla_{x} G\left(x_{\tau}(T)\right),\ \partial_{\tau} x_{\tau}(T) \big\rangle.
\end{aligned}
\end{equation}
The rest of the proof focuses on simplifying the expression above.
We will focus three parts of the integrand, which we colored \color{teal} teal \color{black} and \color{violet} violet\color{black}, and \color{orange} orange\color{black}.
First, let's focus on the \color{teal} teal \color{black} portion of the integrand.
Using integration by parts,
\begin{align}
    \color{teal} \int_{0}^{T} \left\langle \lambda_{\tau}(t),\ \dot{\partial_{\tau} x_{\tau}}(t) \right\rangle\ dt \color{black} &= \left\langle \lambda_{\tau}(T),\ \partial_{\tau} x_{\tau}(T) \right\rangle - \left\langle \lambda_{\tau}(0),\ \partial_{\tau} x_{\tau}(0) \right\rangle - \int_{0}^{T} \left\langle \dot{\lambda}_{\tau}(t),\ \partial_{\tau} x_{\tau}(t) \right\rangle\ dt \nonumber \\
    &= \left\langle \lambda_{\tau}(T),\ \partial_{\tau} x_{\tau}(T) \right\rangle - \int_{0}^{T} \left\langle \dot{\lambda}_{\tau}(t),\ \partial_{\tau} x_{\tau}(t) \right\rangle\ dt. \label{eq:Teal:Tau}
\end{align}
The second line here follows from the fact that $x_{\tau}(0) = X_0(0, \phi)$.
Since $X_0$ does not depend on $\tau$, $\partial_{\tau} x_{\tau}(0) = \partial_{\tau} X_0(0, \phi) = 0$.
Second, let's focus to the \color{violet} violet \color{black} portion of the integrand.
Notice that
\begin{align}
    \color{violet} \int_{0}^{T} &\color{violet}\left\langle \lambda_{\tau}(t),\ \left[ \partial_{y} F(t) \right] \partial_{\tau} x_{\tau}(t - \tau) \right\rangle\ dt \color{black} \nonumber \\
    &= \color{blue}\int_{0}^{\tau} \left\langle \lambda_{\tau}(t),\ \left[ \partial_{y} F(t) \right] \partial_{\tau} x_{\tau}(t - \tau) \right\rangle\ dt \color{black} +  \color{red} \int_{\tau}^{T} \left\langle \lambda_{\tau}(t),\ \left[ \partial_{y} F(t) \right] \partial_{\tau} x_{\tau}(t - \tau) \right\rangle\ dt \color{black} \nonumber \\
    &= \int_{0}^{T - \tau} \left\langle \lambda_{\tau}(t + \tau),\ \left[ \partial_{y} F(t + \tau) \right] \partial_{\tau} x_{\tau}(t) \right\rangle\ dt \nonumber \\
    &= \int_{0}^{T} \mathbb{1}_{t < T - \tau}(t) \left\langle \lambda_{\tau}(t + \tau),\ \left[ \partial_{y} F(t + \tau) \right] \partial_{\tau} x_{\tau}(t) \right\rangle\ dt \nonumber \\
    &= \int_{0}^{T} \mathbb{1}_{t < T - \tau}(t) \left\langle \left[ \partial_y F(t + \tau) \right]^T \lambda_{\tau} (t + \tau),\ \partial_{\tau} x_{\tau}(t) \right\rangle\ dt. \label{eq:Violet:Tau}
\end{align}
To get from the second to third lines of this equation, observe that for $t \in [0, \tau]$, $x_{\tau}(t - \tau) = X_0(t - \tau, \phi)$. 
Since $X_0$ does not depend on $\tau$, we must have 
$$\partial_{\tau} x_{\tau}(t - \tau) = \partial_{\tau} X_0(t - \tau, \phi) = 0 \qquad t \in [0, \tau].$$
Therefore, the \color{blue} blue \color{black} integral is zero.
To get from the \color{red} red \color{black} integral to the third line, we redefine $t$ as $t - \tau$.
Finally, in the last two lines, $\mathbb{1}_{t < T - \tau}(t)$ is the indicator function of the set $(-\infty, T - \tau]$.

\bigskip

Third, since $x_{\tau}(t - \tau) = X_0(t - \tau, \phi)$ for $t < \tau$, the \color{orange} orange \color{black} portion of the integrand is zero in $[0, \tau]$.
Substituting equations \eqref{eq:Teal:Tau} and \eqref{eq:Violet:Tau} into equation \eqref{eq:L:Grad:Tau:Initial} gives
\begin{equation}
\label{eq:L:Grad:Tau:Messy}
\begin{aligned}
    D_{\tau} L\left( \lambda_{\tau}, \theta, \tau, \phi \right) &= \int_{0}^{T} \Big\langle \nabla_{x} \ell\left(x_{\tau}(t) \right) - \dot{\lambda}_{\tau}(t) - \left[ \partial_{x} F(t) \right]^T \lambda_{\tau}(t) - \\
    &\qquad\qquad \mathbb{1}_{t < T - \tau}(t) \left[ F_{y}(t + \tau) \right]^T \lambda_{\tau}(t + \tau),\ \partial_{\tau} x_{\tau}(t) \Big\rangle\ dt \\
    &+ \big\langle \nabla_{x} G\left(x_{\tau}(T)\right) + \lambda_{\tau}(T),\ \partial_{\tau} x_{\tau}(T) \big\rangle \\
    & + \color{orange}\int_{\tau}^{T} \left\langle \lambda_{\tau}(t),\ \left[\partial_{y} F(t) \right] \dot{x}_{\tau}(t - \tau) \right\rangle\ dt\color{black} -\int_{0}^{T} \left\langle \lambda_{\tau}(t),\ \partial_{\tau} F(t) \right\rangle\ dt. 
\end{aligned}
\end{equation}
Thus, if we select $\lambda_{\tau}$ to satisfy the adjoint equation, equation \eqref{eq:adjoint}, then equation \eqref{eq:L:Grad:Tau:Messy} becomes (after a change of variables in the \color{orange} orange \color{black} integral) that
\begin{equation}
\label{eq:L:Grad:Tau}
\begin{aligned}
    \frac{\partial}{\partial \tau} \mathcal{L}\left( x_{\tau} \right) = D_{\tau} L\left( \lambda_{\tau}, \theta, \tau, \phi \right) = &\int_{0}^{T - \tau} \left\langle \left[ \partial_{y} F(t + \tau) \right]^T \lambda_{\tau}(t + \tau),\ \dot{x}_{\tau}(t)\right\rangle \ dt \\
    -&\int_{0}^{T} \left\langle \lambda_{\tau}(t),\ \partial_{\tau} F(t) \right\rangle\ dt.
\end{aligned}
\end{equation}
This concludes the $\tau$ portion of the proof of Theorem \ref{Theorem:1}.

\subsubsection{\texorpdfstring{$\nabla_{\phi} \mathcal{L}$}{}}
\label{Sec:Proof:Grad:Phi}

We now turn to the gradient of $\mathcal{L}$ with respect to $\phi$. 
Once again, fix $(\theta, \tau, \phi) \in \mathbb{R}^p \times (0, \infty) \times \mathbb{R}^q$. 
We denote the solution to equation \eqref{eq:IVP}, $x$, by $x_{\phi}$ to make its dependence on $\phi$ apparent.
We will also allow $\lambda$ to depend on $\phi$.
To make this dependence clear, we write $\lambda_{\phi}$ in place of $\lambda$.
Finally, we restrict our attention to $\lambda_{\phi}$ for which the map $(t, \phi) \to \lambda_{\tau}(t)$ is of class $C^1$.

\bigskip 

In this section, we show that $\mathcal{L}$ is differentiable with respect to $\phi$ and derive an expression for $\nabla_{\phi} \mathcal{L}\left( x_{\phi} \right)$ in terms of the adjoint, equation \eqref{eq:adjoint}.
To begin, we know that 
$$\nabla_{\phi} \mathcal{L}\left( x_{\phi} \right) = \sum_{i = 1}^{q} \frac{\partial}{\partial \phi_{i}} \mathcal{L} \left( x_{\phi} \right).$$
Therefore, suffices to consider the case when $\phi \in \mathbb{R}$ ($q = 1$); we can extend to arbitrary $\phi \in \mathbb{R}^q$ by applying $\mathbb{R}$ case to each component of $\phi$.
Thankfully, we can reuse most of the argument we used to derive the $\tau$ gradient.
By assumption, the map $(t, \phi) \to x_{\phi}(t)$ is of class $C^2$. 
Since $G$ and $\ell$ are of class $C^1$, the chain rule tells us that the maps $\phi \to G(x_{\phi}(T))$ and $(t, \phi) \to \ell(x_{\phi}(t))$ are of class $C^1$ with
$$\frac{\partial}{\partial \phi} G\left(x_{\phi}(T)\right) = \left\langle \nabla_{x} G\left( x_{\phi}(T) \right),\ \partial_{\phi} x_{\phi}(T) \right\rangle,$$
and
$$\frac{\partial}{\partial \phi} \ell\left(x_{\phi}(t)\right) = \left\langle \nabla_{x} \ell\left( x_{\phi}((t) \right),\ \partial_{\phi} x_{\phi}(t) \right\rangle.$$
Using the same argument we used to derive the $\tau$ gradient, $\mathcal{L}$ is differentiable with respect to $\phi$ with 
\begin{equation}
\label{eq:Loss:Grad:Phi}
    \frac{\partial}{\partial \phi} \mathcal{L}\left( x_{\phi} \right) = \left\langle \nabla_{x} G\left( x_{\phi}(T) \right),\ \partial_{\phi} x_{\phi}(T) \right\rangle + \int_{0}^{T} \left\langle \nabla_{x} \ell\left( x_{\phi}((t) \right),\ \partial_{\phi} x_{\phi}(t) \right\rangle\ dt.
\end{equation}
We can now turn our attention to the Lagrangian, equation \eqref{eq:Lagrangian}.
Let $t \in [0, T]$.
Using the same logic we used to derive the $\tau$ gradient, we can conclude that the map $\phi \to F\left( x_{\phi}(t), x_{\phi}(t - \tau), \tau, t, \theta \right)$ is differentiable with 
\begin{equation}
\label{eq:F:Grad:Phi}
    D_{\phi} F(t) = \left[ \partial_{x} F(t) \right] \partial_{\phi} x_{\phi}(t) + \left[ \partial_{y} F(t) \right] \partial_{\phi} x_{\phi}(t - \tau).
\end{equation}
Here, once again, we have adopted the shorthand $F(t)$ as an abbreviation for $F(x_{\phi}(t), x_{\phi}(t - \tau), \tau, t, \theta)$.
Using the same logic we used to derive the $\tau$ gradient, we can conclude that the map $\phi \to \left\langle \lambda_{\phi}(t),\ \dot{x}_{\phi}(t) - F(t) \right\rangle$ is differentiable with 
\begin{equation}
\label{eq:Integrand:Grad:Phi}
    \frac{\partial}{\partial \phi} \left\langle \lambda_{\phi}(t),\ \dot{x}_{\phi}(t) - F(t) \right\rangle = \left\langle \lambda_{\phi}(t),\ \dot{\partial_{\phi} x_{\phi}}(t) - \left[ \partial_{x} F(t) \right] \partial_{\phi} x_{\phi}(t) - \left[ \partial_{y} F(t) \right] \partial_{\phi} x_{\phi}(t - \tau) \right\rangle.
\end{equation}
Since we chose $t$ arbitrarily, this must hold for each $t \in [0, T]$. 
By inspection, the map $t \to \frac{\partial}{\partial \phi} \left\langle \lambda_{\phi}(t),\ \dot{x}_{\phi}(t) - F(t) \right\rangle$ is continuous.
Therefore, the integral portion of the Lagrangian is differentiable with
\begin{equation}
\label{eq:Integral:Grad:Phi}
\begin{aligned}
    \frac{\partial}{\partial \phi} \int_{0}^{T} \Big\langle \lambda_{\phi}(t),\ &\dot{x}_{\phi}(t) - F(t) \Big\rangle\ dt \\
    = &\int_{0}^{T} \left\langle \lambda_{\phi}(t),\ \dot{\partial_{\phi} x_{\phi}}(t) - \partial_{x} F(t) \partial_{\phi} x_{\phi}(t) - \partial_{y} F(t) \partial_{\phi} x_{\phi}(t - \tau) \right\rangle\ dt.
\end{aligned}
\end{equation}
By combining this with equation \eqref{eq:Loss:Grad:Phi}, we can conclude that the map $\phi \to L\left(\lambda_{\phi}, \theta, \tau, \phi \right)$ is differentiable with 
\begin{equation}
\label{eq:L:Grad:Phi:Initial}
\begin{aligned}
    D_{\phi}L \left( \lambda_{\phi}, \theta, \tau, \phi \right) = \int_{0}^{T} &\left\langle \nabla_{x} \ell\left(x_{\phi}(t)\right),\ \partial_{\phi} x_{\phi}(t) \right\rangle + \color{teal} \left\langle \lambda_{\phi}(t),\ \dot{\partial_{\phi} x_{\phi}}(t)\right\rangle \color{black} \\
    - &\left\langle \lambda_{\phi}(t),\ \left[ \partial_{x} F(t)\right] \partial_{\phi}x_{\phi}(t) + \color{violet}\left[ \partial_{y} F(t) \right] \partial_{\phi} x_{\phi}(t - \tau) \color{black} \right\rangle\ dt \\
    + &\left\langle \nabla_{x} G\left( x_{\phi}(T) \right),\ \partial_{\phi} x_{\phi}(T) \right\rangle.
\end{aligned}
\end{equation}
The rest of the proof focuses on simplifying \eqref{eq:L:Grad:Phi:Initial}.
To do this, we focus on two parts of the integral, which we colored \color{teal} teal \color{black} and \color{violet} violet\color{black}.
We treat these portions of the integral similar to how we treated the corresponding parts of the integral in equation \eqref{eq:L:Grad:Tau:Initial}, but with a few important caveats.
To begin, let's focus on the \color{teal} teal \color{black} portion of the integral.
Using integration by parts,
\begin{equation}
\label{eq:Teal:Phi}
\begin{aligned}
    \color{teal} \int_{0}^{T} \left\langle \lambda_{\phi}(t),\ \dot{\partial_{\phi} x_{\phi}}(t) \right\rangle\ dt \color{black} = &\left\langle \lambda_{\phi}(T),\ \partial_{\phi} x_{\phi}(T) \right\rangle - \left\langle \lambda_{\phi}(0),\ \partial_{\phi}X_0(0, \phi) \right\rangle\\
    - &\int_{0}^{T} \left\langle \dot{\lambda}_{\phi}(t),\ \partial_{\phi} x_{\phi}(t) \right\rangle\ dt.
\end{aligned}
\end{equation}
Next, let's focus on the \color{violet} violet \color{black} portion of the integral.
Since $x(t) = X_0(t, \phi)$ for $t \in [-\tau, 0]$, we must have
\begin{align}
    \color{violet} \int_{0}^{T} \Big\langle \lambda_{\phi}(t),\ &\color{violet}\big[ \partial_{y} F(t) \big] \partial_{\phi} x_{\phi}(t - \tau) \Big\rangle\ dt \color{black} \nonumber \\
    = &\color{blue}\int_{0}^{\tau} \Big\langle \lambda_{\phi}(t),\ \left[ \partial_{y} F(t) \right] \partial_{\phi} x_{\phi}(t - \tau) \Big\rangle \ dt \color{black} +  \color{red} \int_{\tau}^{T} \Big\langle \lambda_{\phi}(t),\ \left[ \partial_{y} F(t) \right] \partial_{\phi} x_{\phi}(t - \tau) \Big\rangle\ dt \color{black} \nonumber \\
    = &\int_{0}^{\tau} \Big\langle \lambda_{\phi}(t),\ \left[ \partial_{y} F(t) \right] \partial_{\phi} X_0(t - \tau, \phi) \Big\rangle\ dt \nonumber \\
    + &\int_{0}^{T} \mathbb{1}_{t < T - \tau}(t) \left\langle \left[ \partial_y F(t + \tau) \right]^T \lambda_{\phi} (t + \tau),\ \partial_{\phi} x_{\phi}(t) \right\rangle\ dt. \label{eq:Violet:Phi}
\end{align}
Here, we reduced the \color{red} red \color{black} integral using the same steps we used for the \color{red} red \color{black} integral in the $\tau$ part of the proof.
Substituting equations \eqref{eq:Teal:Phi} and \eqref{eq:Violet:Phi} into equation \eqref{eq:L:Grad:Phi:Initial} gives
\begin{equation}
\label{eq:L:Grad:Phi:Messy}
\begin{aligned}
    D_{\phi} L\left( \lambda_{\phi}, \theta, \tau, \phi \right) = &\int_{0}^{T} \Big\langle \nabla_{x} \ell\left( x_{\phi}(t) \right) - \dot{\lambda}_{\phi}(t) - \left[ \partial_{x} F(t) \right]^T \lambda_{\phi}(t) - \\
    &\qquad\qquad \mathbb{1}_{t < T - \tau}(t) \left[ F_{y}(t + \tau) \right]^T \lambda_{\phi}(t + \tau),\ \partial_{\phi} x_{\phi}(t) \Big\rangle\ dt \\
    - &\int_{0}^{\tau} \Big\langle \lambda_{\phi}(t),\ \left[ \partial_{y} F(t) \right] \partial_{\phi} X_0(t - \tau, \phi) \Big\rangle\ dt \\
    + &\Big\langle \nabla_{x} G\left( x_{\phi}(T) \right) + \lambda_{\phi}(T),\ \partial_{\phi} x_{\phi}(T) \Big\rangle - \Big\langle \lambda_{\phi}(0),\ \partial_{\phi} X_0(0, \phi) \Big\rangle.
\end{aligned}
\end{equation}
Thus, if we select $\lambda_{\phi}$ to satisfy the adjoint equation, equation \eqref{eq:adjoint}, then equation \eqref{eq:L:Grad:Phi:Messy} simplifies to the following:
\begin{equation}
\label{eq:L:Grad:Phi:Scalar}
    D_{\phi} L\left( \lambda_{\phi}, \theta, \tau, \phi \right) = -\int_{0}^{\tau} \Big\langle \lambda_{\phi}(t),\ \left[ \partial_{y} F(t) \right] \partial_{\phi} X_0(t - \tau, \phi) \Big\rangle\ dt - \Big\langle \lambda_{\phi}(0),\ \partial_{\phi} X_0(0, \phi) \Big\rangle
\end{equation}
We can now generalize this result to vector-valued $\phi$.
In particular, notice that if we replace $\phi$ with $\phi_i$, then the right-hand side of \eqref{eq:L:Grad:Phi:Scalar} is the $i$'th component of
$$-\int_{0}^{\tau} \left[ \partial_{\phi} X_0(t - \tau, \phi) \right]^T \left[ \partial_{y} F(t) \right]^T \lambda_{\phi}(t)\ dt - \left[ \partial_{\phi} X_0(0, \phi) \right]^T \lambda_{\phi}(0).$$
In particular, this means that
\begin{equation}
\label{eq:L:Grad:Phi}
\begin{aligned}
    \nabla_{\phi}\mathcal{L} \left( x_{\phi}\right) = D_{\phi} L \left( \lambda_{\phi}, \theta, \tau, \phi \right) = -\int_{0}^{\tau} &\left[ \partial_{\phi} X_0(t - \tau, \phi) \right]^T \left[ \partial_{y} F(t) \right]^T \lambda_{\phi}(t)\ dt +\\
    - &\left[ \partial_{\phi} X_0(0, \phi) \right]^T \lambda_{\phi}(0).
\end{aligned}
\end{equation}
This concludes the proof of Theorem \ref{Theorem:1}.
\section{Data Tables}
\label{Sec:Tables}

\begin{table}[hbt]
    \centering
    \caption{Results for the Delay Exponential Decay Equation for noise level $0.1$}
    \rowcolors{2}{white}{olive!20!green!15}

    \begin{tabulary}{\linewidth}{C C C C}
        \toprule[0.3ex]
        Parameter &  True Value & Mean Learned Value & Standard Deviation \\
        \midrule[0.1ex]
        $\theta_0$ & $-2.0$ & $-2.0324$ & $0.0291$ \\
        $\theta_1$ & $-2.0$ & $-2.0110$ & $0.0204$ \\
        $\tau$ & $1.0$ & $1.0044$ & $0.0068$ \\
        $a$ & $1.5$ & $1.5346$ & $0.0946$ \\
        $b$ & $4.0$ & $4.0071$ & $0.0683$ \\
        \bottomrule[0.3ex]    
    \end{tabulary}
        
    \label{table:Exponential:0.1}
\end{table}

\begin{table}[hbt]
    \centering
    \caption{Results for the Delay Exponential Decay Equation for noise level $0.9$}
    \rowcolors{2}{white}{olive!20!green!15}

    \begin{tabulary}{\linewidth}{C C C C}
        \toprule[0.3ex]
        Parameter &  True Value & Mean Learned Value & Standard Deviation \\
        \midrule[0.1ex]
        $\theta_0$ & $-2.0$ & $-2.0966$ & $0.1827$ \\
        $\theta_1$ & $-2.0$ & $-2.0220$ & $0.3081$ \\
        $\tau$ & $1.0$ & $1.0233$ & $0.0475$ \\
        $a$ & $1.5$ & $1.2415$ & $0.4970$ \\
        $b$ & $4.0$ & $3.5446$ & $0.6803$ \\
        \bottomrule[0.3ex]    
    \end{tabulary}
        
    \label{table:Exponential:0.9}
\end{table}

\begin{table}[hbt]
    \centering
    \caption{Results for the Logistic Decay Equation for noise level $0.1$}
    \rowcolors{2}{olive!20!green!15}{white}

    \begin{tabulary}{\linewidth}{C C C C}
        \toprule[0.3ex]
        Parameter & True Value & Mean Learned Value & Standard Deviation \\
        \midrule[0.1ex]
        $\theta_0$  & $2.0$     & $2.0033$ & $0.0203$ \\
        $\theta_1$  & $1.5$     & $1.5032$ & $0.0174$ \\
        $\tau$      & $1.0$     & $0.9994$ & $0.0039$ \\
        $A$         & $-0.5$    & $0.2484$ & $0.1390$ \\
        $\omega$    & $3.0$     & $5.6099$ & $0.2406$ \\
        $b$         & $2.0$     & $2.3848$ & $0.0619$ \\
        \bottomrule[0.3ex]    
    \end{tabulary}
        
    \label{table:Logistic:0.1}
\end{table}

\begin{table}[hbt]
    \centering
    \caption{Results for the Logistic Delay Equation for noise level $0.9$}
    \rowcolors{2}{olive!20!green!15}{white}

    \begin{tabulary}{\linewidth}{C C C C}
        \toprule[0.3ex]
        Parameter &  True Value & Mean Learned Value & Standard Deviation \\
        \midrule[0.1ex]
        $\theta_0$  & $2.0$     & $2.0611$ & $0.1291$ \\
        $\theta_1$  & $1.5$     & $1.5186$ & $0.1376$ \\
        $\tau$      & $1.0$     & $0.9998$ & $0.0134$ \\
        $A$         & $-0.5$    & $0.4427$ & $1.0127$ \\
        $\omega$    & $3.0$     & $5.7477$ & $0.4873$ \\
        $b$         & $2.0$     & $2.4615$ & $0.4367$ \\
        \bottomrule[0.3ex]    
    \end{tabulary}
        
    \label{table:Logistic:0.9}
\end{table}

\begin{table}[hbt]
    \centering
    \caption{Results for the ENSO for noise level $0.1$}
    \rowcolors{2}{white}{olive!20!green!15}

    \begin{tabulary}{\linewidth}{C C C C}
        \toprule[0.3ex]
        Parameter &  True Value & Mean Learned Value & Standard Deviation \\
        \midrule[0.1ex]
        $\theta_0$  & $1.0$     & $5.1961$  & $0.1089$ \\
        $\theta_1$  & $1.0$     & $0.9360$  & $0.1052$ \\
        $\theta_2$  & $0.75$    & $1.0738$  & $0.0790$ \\
        $\tau$      & $5.0$     & $0.7894$  & $0.0330$ \\
        $A$         & $-0.25$   & $-0.4988$ & $0.1092$ \\
        $\omega$    & $1.0$     & $0.9686$  & $0.0578$ \\
        $b$         & $1.5$     & $1.5400$  & $0.0799$ \\
        \bottomrule[0.3ex]    
    \end{tabulary}
        
    \label{table:ENSO:0.1}
\end{table}

\begin{table}[hbt]
    \centering
    \caption{Results for the ENSO for noise level $0.9$}
    \rowcolors{2}{white}{olive!20!green!15}

    \begin{tabulary}{\linewidth}{C C C C}
        \toprule[0.3ex]
        Parameter &  True Value & Mean Learned Value & Standard Deviation \\
        \midrule[0.1ex]
        $\theta_0$  & $1.0$     & $1.0335$  & $0.3458$ \\
        $\theta_1$  & $1.0$     & $1.2403$  & $0.3205$ \\
        $\theta_2$  & $0.75$    & $0.9000$  & $0.1589$ \\
        $\tau$      & $5.0$     & $5.2560$  & $0.5367$ \\
        $A$         & $-0.25$   & $-0.7483$ & $0.2748$ \\
        $\omega$    & $1.0$     & $1.0548$  & $0.2136$ \\
        $b$         & $1.5$     & $1.5414$  & $0.2937$ \\
        \bottomrule[0.3ex]  
    \end{tabulary}
        
    \label{table:ENSO:0.9}
\end{table}

\begin{table}[hbt!]
    \centering
    \caption{Results for the Cheyne-Stoke Respiration model for noise level $0.1$}
    \rowcolors{2}{olive!20!green!15}{white}

    \begin{tabulary}{\linewidth}{C C C C}
        \toprule[0.3ex]
        Parameter &  True Value & Mean Learned Value & Standard Deviation \\
        \midrule[0.1ex]
        $p$         & $1.0$     & $1.0073$  & $0.0078$ \\
        $V_0$       & $7.0$     & $7.2607$  & $0.0814$ \\
        $\alpha$    & $2.0$     & $2.1257$  & $0.0733$ \\
        $\tau$      & $0.25$    & $0.2489$  & $0.0011$ \\
        $a$         & $-5.0$    & $2.1722$  & $0.1455$ \\
        $b$         & $2.0$     & $1.9633$  & $0.0278$ \\
        \bottomrule[0.3ex]    
    \end{tabulary}
        
    \label{table:Cheyne:0.1}
\end{table}

\begin{table}[hbt!]
    \centering
    \caption{Results for the Cheyne-Stoke Respiration model for noise level $0.9$}
    \rowcolors{2}{olive!20!green!15}{white}

    \begin{tabulary}{\linewidth}{C C C C}
        \toprule[0.3ex]
        Parameter &  True Value & Mean Learned Value & Standard Deviation \\
        \midrule[0.1ex]
        $p$         & $1.0$     & $1.0262$  & $0.0704$ \\
        $V_0$       & $7.0$     & $8.3455$  & $0.8622$ \\
        $\alpha$    & $2.0$     & $2.7853$  & $0.7494$ \\
        $\tau$      & $0.25$    & $0.2439$  & $0.0150$ \\
        $a$         & $-5.0$    & $1.6572$  & $1.3137$ \\
        $b$         & $2.0$     & $1.9135$  & $0.3144$ \\
        \bottomrule[0.3ex]    
    \end{tabulary}
        
    \label{table:Cheyne:0.9}
\end{table}

\begin{table}[hbt!]
    \centering
    \caption{$\phi$ results for the HIV model for noise level $0.1$}
    \rowcolors{2}{white}{olive!20!green!15}

    \begin{tabulary}{\linewidth}{C C C C}
        \toprule[0.3ex]
        Parameter &  True Value & Mean Learned Value & Standard Deviation \\
        \midrule[0.1ex]
        $A[T^*]$            & $1.0$     & $1.0000$  & N/A       \\
        $A[V_I]$            & $3.0$     & $2.0032$  & $0.0073$  \\
        $A[V_{NI}]$         & $0.0$     & $1.0000$  & N/A       \\
        $\omega[T^*]$       & $1.0$     & $1.2000$  & N/A       \\
        $\omega[V_{I}]$     & $3.0$     & $5.2488$  & $0.0140$  \\
        $\omega[V_{NI}]$    & $0.0$     & $0.0000$  & N/A       \\
        $b[T^*]$            & $10.0$    & $9.9144$  & $0.0033$ \\
        $b[V_{I}]$          & $8.0$     & $7.5462$  & $0.0177$ \\
        $b[V_{NI}]$         & $0.0$     & $0.0032$  & $0.0633$ \\
        \bottomrule[0.3ex]    
    \end{tabulary}
        
    \label{table:HIV:Phi:0.1}
\end{table}

\begin{table}[hbt!]
    \centering
    \caption{$\phi$ results for the HIV model for noise level $0.3$}
    \rowcolors{2}{white}{olive!20!green!15}

    \begin{tabulary}{\linewidth}{C C C C}
        \toprule[0.3ex]
        Parameter &  True Value & Mean Learned Value & Standard Deviation \\
        \midrule[0.1ex]
        $A[T^*]$            & $1.0$     & $1.0$     & N/A       \\
        $A[V_I]$            & $3.0$     & $2.0083$  & $0.0117$  \\
        $A[V_{NI}]$         & $0.0$     & $1.0$     & N/A       \\
        $\omega[T^*]$       & $1.0$     & $1.2$     & N/A       \\
        $\omega[V_{I}]$     & $3.0$     & $5.2569$  & $0.0296$  \\
        $\omega[V_{NI}]$    & $0.0$     & $0.0$     & N/A       \\
        $b[T^*]$            & $10.0$    & $9.9186$  & $0.0064$ \\
        $b[V_{I}]$          & $8.0$     & $7.5408$  & $0.0529$ \\
        $b[V_{NI}]$         & $0.0$     & $-0.0804$ & $0.2171$ \\
        \bottomrule[0.3ex]    
    \end{tabulary}
        
    \label{table:HIV:Phi:0.3}
\end{table}

\begin{table}[hbt!]
    \centering
    \caption{$\phi$ results for the HIV model for noise level $0.9$}
    \rowcolors{2}{white}{olive!20!green!15}

    \begin{tabulary}{\linewidth}{C C C C}
        \toprule[0.3ex]
        Parameter &  True Value & Mean Learned Value & Standard Deviation \\
        \midrule[0.1ex]
        $A[T^*]$            & $1.0$     & $1.0$     & N/A       \\
        $A[V_I]$            & $3.0$     & $2.0009$  & $0.0130$  \\
        $A[V_{NI}]$         & $0.0$     & $1.0$     & N/A       \\
        $\omega[T^*]$       & $1.0$     & $1.2$     & N/A       \\
        $\omega[V_{I}]$     & $3.0$     & $5.3153$  & $0.0607$  \\
        $\omega[V_{NI}]$    & $0.0$     & $0.0$     & N/A       \\
        $b[T^*]$            & $10.0$    & $9.9311$  & $0.0148$ \\
        $b[V_{I}]$          & $8.0$     & $7.5066$  & $0.1529$ \\
        $b[V_{NI}]$         & $0.0$     & $0.0167$  & $0.5083$ \\
        \bottomrule[0.3ex]    
    \end{tabulary}
        
    \label{table:HIV:Phi:0.9}
\end{table}

\end{document}